\documentclass[journal]{IEEEtran}

\ifCLASSINFOpdf

\else

\fi

\usepackage{graphicx}
\usepackage{amsmath}
\usepackage{amssymb}
\usepackage{booktabs}
\usepackage{multirow}
\usepackage{times}
\usepackage{setspace}
\usepackage{subfigure}
\usepackage{graphicx}
\usepackage{amsfonts}
\usepackage{comment}
\usepackage{color}
\usepackage{xcolor}
\usepackage{tabularx}
\usepackage{makecell}
\usepackage{enumerate}
\usepackage[T1]{fontenc}
\usepackage{hyperref}
\usepackage{bm}
\usepackage{changepage}
\usepackage{lineno}
\usepackage{array}
\usepackage{dsfont}
\usepackage[numbers]{natbib}
\usepackage[ruled,lined,linesnumbered]{algorithm2e}

\usepackage{ragged2e}
\usepackage{tabularx}
\usepackage{color, colortbl}
\usepackage{pifont}
\usepackage{hhline}

\definecolor{ForestGreen}{rgb}{0.13, 0.55, 0.13}
\definecolor{Maroon}{rgb}{0.69, 0.19, 0.0}
\definecolor{Gray}{gray}{0.8}

\newcommand{\cmark}{\textcolor{ForestGreen}{\ding{51}}}%
\newcommand{\xmark}{\textcolor{Maroon}{\ding{55}}}%

\graphicspath{ {figures/} }

\ifCLASSINFOpdf
\else
\fi

\begin{document}
%\title{Unsupervised Domain Adaptation for Learning Image Emotion Distributions with GANs}
%\title{Continuous Probability Distribution Prediction of Image Emotions via Sparse Multi-Task Regression}
%\title{Adapting Image Emotions with Generative Adversarial Networks}
%\title{CycleEmotionGAN++: Unsupervised Domain Adaptation for Image Emotion Analysis with Generative Adversarial Networks}
%Unsupervised Domain Adaptation for Image Emotion Analysis with Emotional Semantics-Preserved and Feature-Aligned CycleGAN
\title{Emotional Semantics-Preserved and Feature-Aligned CycleGAN for Visual Emotion Adaptation}

%\author{Sicheng~Zhao$^{\dagger}$,\IEEEmembership{~Senior~Member,~IEEE},~Xuanbai~Chen$^{\dagger}$,~Xiangyu~Yue,~Chuang~Lin,~Pengfei~Xu,\\~Ravi~Krishna,~Kurt~Keutzer,\IEEEmembership{~Fellow,~IEEE}
\author{Sicheng~Zhao,\IEEEmembership{~Senior~Member,~IEEE},~Xuanbai~Chen,~Xiangyu~Yue,~Chuang~Lin,~Pengfei~Xu,~Ravi~Krishna,\\~Jufeng~Yang,~Guiguang~Ding,~Alberto~L.~Sangiovanni-Vincentelli,~\IEEEmembership{Fellow, IEEE},~Kurt~Keutzer,\IEEEmembership{~Life~Fellow,~IEEE}
%\thanks{Copyright (c) 2013 IEEE. Personal use of this material is permitted. However, permission to use this material for any other purposes must be obtained from the IEEE by sending a request to pubs-permissions@ieee.org.}
\thanks{Manuscript received August 4, 2020; revised November 24, 2020. This work was supported by the National Natural Science Foundation of China (No. 61701273) and Berkeley DeepDrive. Sicheng Zhao and Xuanbai Chen contributed equally. Corresponding author: Sicheng Zhao.}
\thanks{S. Zhao and G. Ding are with the School of Software, Tsinghua University, Beijing 100084, China. (e-mail: schzhao@gmail.com, dinggg@tsinghua.edu.cn).}
\thanks{X. Chen and J. Yang are with College of Computer Science, Nankai University, Tianjin 300350, China (e-mail: chenxuanbai@126.com, yangjufeng@nankai.edu.cn).}
\thanks{X. Yue, R. Krishna, A. L. Sangiovanni-Vincentelli, and K. Keutzer are with the Department of Electrical Engineering and Computer Sciences, University of California, Berkeley, 94709 USA (e-mail: xyyue@berkeley.edu, ravi.krishna@berkeley.edu, alberto@berkeley.edu, keutzer@berkeley.edu).}
\thanks{C. Lin and P. Xu are with Didi Chuxing, Beijing, China (e-mail: chuanglin.hit@outlook.com, xupengfeipf@didiglobal.com).}
}

\markboth{Under Review}
{Shell Zhao{\textit{et al.}}: Adapting Image Emotions with Generative Adversarial Networks}

\maketitle
\begin{abstract}
Thanks to large-scale labeled training data, deep neural networks (DNNs) have obtained remarkable success in many vision and multimedia tasks. However, because of the presence of domain shift, the learned knowledge of the well-trained DNNs cannot be well generalized to new domains or datasets that have few labels. Unsupervised domain adaptation (UDA) studies the problem of transferring models trained on one labeled source domain to another unlabeled target domain. In this paper, we focus on UDA in visual emotion analysis for both emotion distribution learning and dominant emotion classification. Specifically, we design a novel end-to-end cycle-consistent adversarial model, termed CycleEmotionGAN++. First, we generate an adapted domain to align the source and target domains on the pixel-level by improving CycleGAN with a multi-scale structured cycle-consistency loss. During the image translation, we propose a dynamic emotional semantic consistency loss to preserve the emotion labels of the source images. Second, we train a transferable task classifier on the adapted domain with feature-level alignment between the adapted and target domains. We conduct extensive UDA experiments on the Flickr-LDL \& Twitter-LDL datasets for distribution learning and ArtPhoto \& FI datasets for emotion classification. The results demonstrate the significant improvements yielded by the proposed CycleEmotionGAN++ as compared to state-of-the-art UDA approaches.
\end{abstract}
%The results demonstrate that the proposed CycleEmotionGAN++ achieves superior performance as compared to the state-of-the-art UDA approaches.

%, the labels of which are difficult to obtain
%The emotion labels of the source images are simultaneously preserved by the proposed dynamic emotional semantic consistency.


\IEEEpeerreviewmaketitle

\section{Introduction}
\label{sec:Introduction}

It has been revealed that visual content such as images and videos can evoke strong emotions for human beings~\cite{lang1979bio}.
With the popularization of various mobile devices, cameras, and the Internet, people have become accustomed to recording their activities, sharing their experiences, and expressing their opinions by using images and videos appearing alongside text in social networks~\cite{zhao2017real}. The generation of a large amount of multimedia data has made it convenient for researchers to process and analyze visual content. Understanding the implied emotions in the data is of great importance to behavior sciences and enables various applications including blog recommendation, decision making, and appreciation of art~\cite{borth2013large}.

\begin{figure}[!t]
    \centering
        \subfigure[Train on Twitter-LDL, test on Twitter-LDL]
        {\includegraphics[width=2cm]{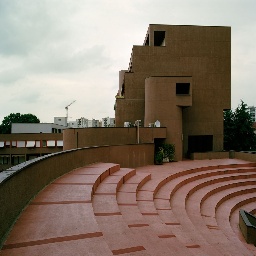}
        \quad
        \includegraphics[width=2.85cm]{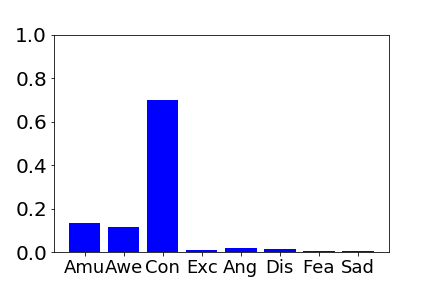}
        \includegraphics[width=2.85cm]{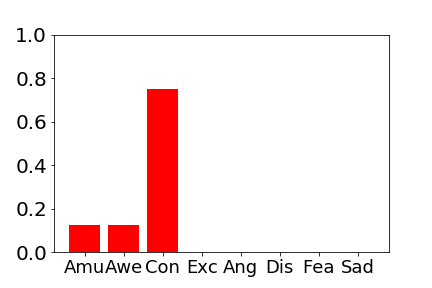}}
        
        \subfigure[Train on Twitter-LDL, test on Flickr-LDL]
        {\includegraphics[width=2cm]{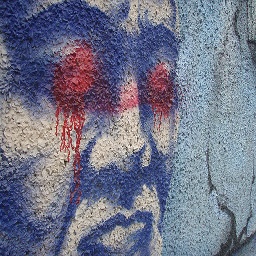}
        \quad
        \includegraphics[width=2.85cm]{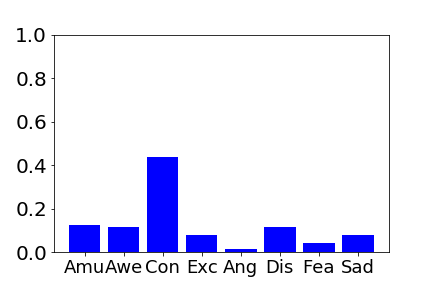}
        \includegraphics[width=2.85cm]{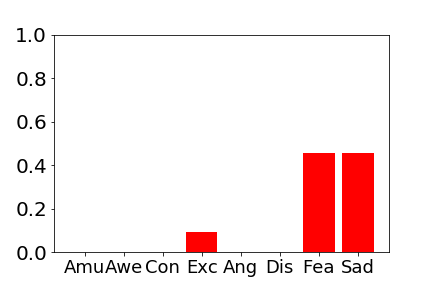}}
        
        \caption{An example of domain shift when performing the emotion distribution learning task. The classifier trained on Twitter-LDL is tested on the top image from Twitter-LDL and the bottom image from Flickr-LDL. The objects displayed, from left to right, are: the original image, the predicted emotion distribution, and the ground truth distribution.}
    \label{fig:domain_shift1}
\end{figure}

\begin{figure}[!t]
    \centering
    \includegraphics[width=0.95\linewidth]{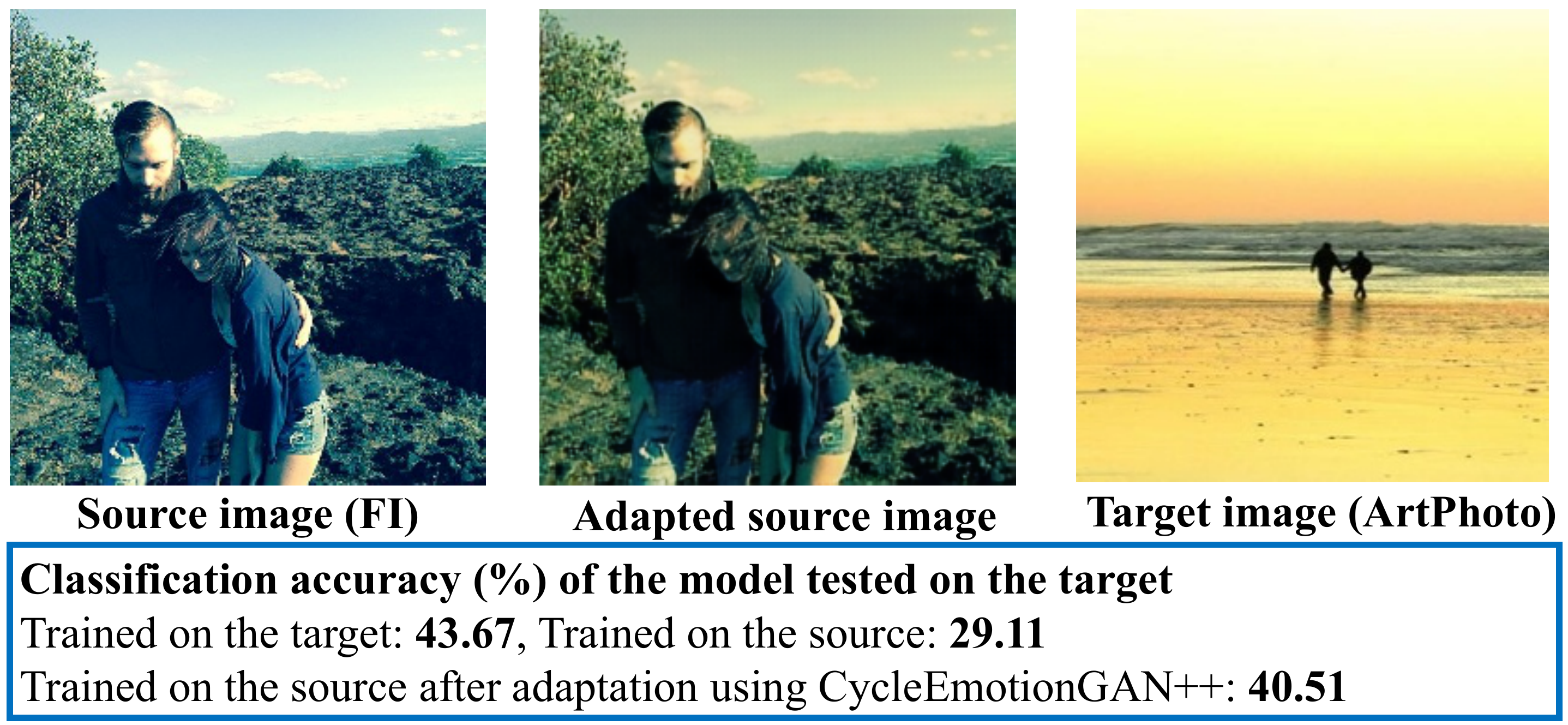} 
    \caption{An example of domain shift when performing dominant emotion classification. The overall accuracy of a state-of-the-art model (\citeauthor{he2016deep}~\cite{he2016deep}) drops from 43.67$\%$ (trained on the target ArtPhoto) to 29.11$\%$ (trained only on the source FI). We propose CycleEmotionGAN++, which achieves significant performance improvements (11.40\%) over the source-trained model baselines.}
    \label{fig:domain_shift2}
\end{figure}

Recognizing the emotions induced by image content is often referred to as visual emotion analysis (VEA)~\cite{zhao2019pdanet}. This task mainly faces two challenges: the affective gap~\cite{zhao2014exploring} and perception subjectivity~\cite{peng2015mixed,zhao2016predicting}. The former one reveals that the extracted feature-level information is inconsistent with the high-level emotions felt by human beings; while the latter indicates that due to different personal and contextual factors such as education background, culture, and personality, different people may produce different emotions after viewing the same image~\cite{yang2017learning,zhao2018affective}. In order to bridge the affective gap, a variety of hand-crafted features have been designed such as color and texture~\cite{machajdik2010affective}, shape~\cite{lu2012shape}, principles-of-art~\cite{zhao2014exploring}, and adjective noun pairs~\cite{borth2013large}. These methods mainly map the image content to one dominant emotion category (DEC).
%More recently, \citeauthor{zhu2017dependency}~\cite{zhu2017dependency} proposed an unified CNN-RNN approach to solve the affective gap problem.
To deal with the subjectivity issue, either personalized emotion perception is predicted for each viewer~\cite{zhao2016predicting}, or an emotion distribution is learned for each image~\cite{yang2017joint,zhao2017approximating,zhao2017learning}.

Recently, convolutional neural networks (CNNs) have been employed to deal with the issue of mapping image content to emotions~\cite{peng2015mixed,zhu2017dependency,yang2017joint,yang2018weakly,zhao2019pdanet}. 
These CNN-based VEA methods can perform well on large-scale training datasets with labels. 
However, due to the presence of \emph{domain shift} or \emph{dataset bias}~\cite{torralba2011unbiased}, the performance of a model directly transferred from one labeled domain to another unlabeled domain drops significantly~\cite{zhao2020review}, as shown in Fig.~\ref{fig:domain_shift1} and Fig.~\ref{fig:domain_shift2}.
%However, due to the \emph{domain shift} or \emph{dataset bias} presented in~\cite{torralba2011unbiased}, a model trained in one labeled domain can not perform well in another unlabeled domain directly. 
%As shown in Fig.~\ref{fig:domain_shift1} and Fig.~\ref{fig:domain_shift2}, the \emph{domain shift} suffer from the variation of scenes, lighting, weather conditions and so on.
%From Fig.~\ref{fig:domain_shift1} and Fig.~\ref{fig:domain_shift2}, we can have a deeper understanding of \emph{domain shift} or \emph{dataset bias}. 
Domain adaptation (DA) is a machine learning paradigm that tries to train a model on a source domain that can perform well on a different, but related, target domain. 
%We generate an adapted domain on which the model trained performs well in the target domain. 
To the best of our knowledge, although DA has been used in various vision tasks~\cite{patel2015visual,zhao2020review}, it has rarely been used for VEA.

In this paper, we study the unsupervised domain adaptation (UDA) problem of analyzing visual emotions in one labeled source domain and adapting it to another unlabeled target domain. A novel cycle-consistent adversarial UDA model termed CycleEmotionGAN++ is proposed for dominant emotion classification and emotion distribution learning. First, we generate an intermediate domain to align the source and target domains on the pixel-level based on generative adversarial network (GAN)~\cite{goodfellow2014generative}. Since this mapping from the source domain to the intermediate domain is highly under-constrained~\cite{zhu2017unpaired}, we couple an inverse mapping and a cycle-consistency loss to enforce the reconstructed source to be as similar as possible to the original source. We add a multi-scale structural similarity loss to the original cycle-consistency loss to better preserve the high-frequency detailed information, and this combination is defined as the mixed cycle-consistency loss. In addition, we complement the mixed CycleGAN loss with a dynamic emotional semantic consistency (DESC) loss that penalizes large semantic changes between the adapted and source images. Two different classifiers are trained on the source domain and adapted domain respectively to dynamically preserve the semantic information. In order to make the adapted domain and target domain as similar as possible, we also add feature-level alignment by training a discriminator to maximize the probability of correctly classifying feature maps from adapted images and target images. In this way, the CycleEmotionGAN++ model can adapt the source domain images to appear as if they were drawn from the target domain. Eventually, by optimizing the mixed CycleGAN loss, DESC loss, feature-level alignment loss, and task classification loss alternately, a transferable CycleEmotionGAN++ model is learned.

\begin{table*}
\centering\footnotesize
\caption{Comparison of the proposed CycleEmotionGAN++ model with several state-of-the-art domain adaptation methods. The full names of each attribute from the second to the last column are pixel-level alignment, feature-level alignment, emotional semantic consistency, cycle-consistency, multi-scale structural similarity, emotion distribution learning task, and dominant emotion classification task, respectively.}    
\begin{center}
\begin{tabular}{c|c c c c c c c}
%{p{4.5cm}|p{1.5cm} p{1.5cm} p{1.5cm} p{1.5cm} p{1.5cm} p{1.5cm} p{1.5cm}}
\toprule
Method & pixel & feature & semantic & cycle & msssim & distrib & classifi\\
\hline  
CycleGAN~\cite{zhu2017unpaired} & \cmark & \xmark & \xmark & \cmark & \xmark & \xmark & \xmark \\
SAPE~\cite{yan2016automatic} & \cmark & \cmark & \xmark & \xmark & \xmark & \xmark & \xmark \\
EICT~\cite{liu2018emotional} & \cmark & \cmark & \xmark & \xmark & \xmark & \xmark & \xmark \\
TAECT~\cite{liu2018texture} & \xmark & \cmark & \xmark & \xmark & \xmark & \xmark & \xmark \\
ADDA~\cite{tzeng2017adversarial} & \xmark & \cmark & \textbf{static} & \xmark & \xmark & \xmark & \xmark\\
SimGAN~\cite{shrivastava2017learning} & \cmark & \xmark & \xmark & \xmark & \xmark & \xmark & \xmark\\
CyCADA~\cite{hoffman2018CyCADA} & \cmark & \cmark & \textbf{static} & \cmark & \xmark & \xmark & \xmark\\
EmotionGAN~\cite{zhao2018emotiongan} & \cmark & \xmark & \textbf{static} & \xmark & \xmark & \cmark & \xmark\\
\hline
CycleEmotionGAN (Ours) & \cmark & \xmark & \textbf{static} & \cmark& \xmark & \xmark & \cmark\\ 
\textbf{CycleEmotionGAN++ (ours)} & \cmark & \cmark & \textbf{dynamic} & \cmark & \cmark & \cmark & \cmark\\
\bottomrule
\end{tabular}  
\label{tab:compare}
\end{center} 
\end{table*}

In summary, the contributions of this paper are threefold:
\begin{itemize}
    \item We propose to adapt visual emotions from one source domain to another target domain by using a novel end-to-end cycle-consistent adversarial model. To the best of our knowledge, this is the first work on unsupervised domain adaptation for both emotion distribution learning and dominant emotion classification tasks.
    \item We develop a novel adversarial model, termed CycleEmotionGAN++, by alternately optimizing the mixed CycleGAN loss, DESC loss, feature-level alignment loss, and task classification loss. The adapted images are indistinguishable from the target images, thanks to the mixed CycleGAN loss which can preserve the contrast and detailed information better by adding the multi-scale structural similarity, the DESC loss which can preserve the annotation information of the source images, and the feature-level alignment loss which can align adapted and target images on the feature-level.
    \item We conduct extensive experiments on four datasets: Twitter-LDL and Flickr-LDL for emotion distribution learning, and ArtPhoto and FI for dominant emotion classification. The results demonstrate the significant improvements yielded by CycleEmotionGAN++.
\end{itemize}

CycleEmotionGAN++ is extended from CycleEmotionGAN, which was previously introduced in our AAAI 2019 paper~\cite{zhao2019cycleemotiongan}. The improvements include the following three aspects. First, the image translation is conducted with mixed CycleGAN by enforcing the multi-scale structural similarity and with dynamic emotional semantic consistency; feature-level alignment is added to better align the source and target domains. Second, we conduct more UDA experiments for both emotion distribution learning and dominant emotion classification. Third, we provide a more comprehensive review to introduce the background and comparison.

\section{Related Work}
\label{sec:RelatedWork}

\noindent\textbf{Emotion Representation.} Two models are typically employed by psychologists to represent emotions: categorical emotion states (CES) and dimensional emotion space (DES)~\cite{zhao2018affective}. CES models usually consider classifying emotions into several basic categories, such as Mikels' eight emotions (\emph{amusement}, \emph{anger}, \emph{awe}, \emph{contentment}, \emph{disgust}, \emph{excitement}, \emph{fear}, \emph{sadness}). DES models usually employ a Cartesian space to represent emotions, such as the 3D valence-arousal-dominance (VAD) space~\cite{lang2005international}. CES is intuitive for human beings to understand in labeling emotions for images, while DES is more abstract and fine-grained. In this paper, the classic Mikels' eight emotions are employed as our emotion model. 

\noindent\textbf{Visual Emotion Analysis.} Similar to other machine learning and computer vision tasks, visual emotion analysis (VEA) also involves feature extraction and classifier training~\cite{zhao2017continuous,zhao2018affective}. While classifier training is mainly based on existing machine learning algorithms, the main focus in VEA is extracting discriminative features. In the early stage, researchers mainly hand-crafted features on different levels~\cite{zhao2015predicting}, including low-level features such as color and texture~\cite{machajdik2010affective}, shape~\cite{lu2012shape}; mid-level features such as principles-of-art~\cite{zhao2014exploring}, composition~\cite{machajdik2010affective}, and attributes~\cite{yuan2013sentribute}; and high-level features such as skins~\cite{machajdik2010affective} and adjective noun pairs~\cite{borth2013large, chen2014object}. Some other methods try to combine various features on different levels~\cite{zhao2017approximating,zhao2018predicting}. A learning-based visual
affective filtering framework has been proposed to synthesize user-specified emotions onto arbitrary input images or videos~\cite{li2015data}.

In recent years, CNNs have made a great success on many machine learning tasks including VEA. \citeauthor{you2016building}~\cite{you2016building} built a large dataset for VEA and designed a progressive CNN architecture to make use of noisily labeled data for sentiment polarity classification. Various methods have been proposed to predict the probability distributions of image emotions~\cite{yang2017learning, yang2017joint, zhao2018emotiongan, zhao2017approximating, zhao2017learning, zhao2018discrete}. In order to predict emotion distributions more rapidly and accurately, some methods~\cite{peng2015mixed,zhao2018emotiongan} fine-tune CNN models pretrained on ImageNet. \citeauthor{you2015robust}~\cite{you2015robust} and \citeauthor{zhao2019cycleemotiongan}~\cite{zhao2019cycleemotiongan} also fine-tuned a pre-trained CNN to classify visual emotions on a new large-scale Flickr and Instagram dataset~\cite{you2016building} respectively. \citeauthor{yang2018weakly}~\cite{yang2018weakly} proposed 
weakly supervised coupled networks (WSCNet) with two branches: sentiment map detection and coupled sentiment classification to improve the classification accuracy. Local information is also considered in~\cite{you2017visual,zhao2019pdanet}. \citeauthor{rao2019learning}~\cite{rao2019learning} learned multi-level deep representations (MldrNet), including aesthetics CNN, AlexNet, and texture CNN. \citeauthor{yang2018retrieving}~\cite{yang2018retrieving} optimized both retrieval and classification losses by using the sentiment constraints adapted from the triplet constraints, which is able to explore the hierarchical relation of emotion labels.

The above mapping methods between image content and emotions are all performed in a supervised manner. Please refer to~\cite{zhao2018affective} for a comprehensive survey on VEA. In this paper, we study how to adapt the models trained from one labeled source domain to another unlabeled target domain for VEA.

\noindent\textbf{Unsupervised Domain Adaptation (UDA).}
In the early years, domain adaptation was introduced in a transform-based adaptation technique for object recognition~\cite{saenko2010adapting}.  \citeauthor{torralba2011unbiased}~\cite{torralba2011unbiased} conducted a comparison study using a set of popular datasets and conducted a deep discussion regarding dataset bias. For UDA, it has been explored in~\cite{patel2015visual} for image classification task with extensive reviews of some non-deep approaches. These methods mainly focused on feature space alignment through minimizing the distance between the source domain and target domain, either by sample re-weighting techniques~\cite{gong2013connecting,huang2007correcting} or by constructing intermediate subspace representations~\cite{fernando2013unsupervised, gong2012geodesic}. 

Recent efforts have shifted to employing deep models. In order to represent the source and target domains, \citeauthor{zhuo2017deep}~\cite{zhuo2017deep} proposed a conjoined architecture with two streams for UDA. Labeled source data is used for the supervised task loss and deep UDA models are usually trained jointly with another loss, such as a discrepancy loss, adversarial loss, or self-supervision loss, to deal with domain shift.

Discrepancy-based methods mainly measure the discrepancy directly between the source and target domains on corresponding activation layers, such as the multiple kernel variant of maximum mean discrepancies on the fully connected (FL) layers~\cite{long2015learning}, correlation alignment (CORAL)~\cite{sun2017correlation} and geodesic correlation alignment~\cite{wu2019squeezesegv2} on the last FL layer, CORAL on both the last conv layer and FL layer~\cite{zhuo2017deep}, and contrastive domain discrepancy on multiple FL layers~\cite{kang2019contrastive}. 
Adversarial discriminative models usually employ an adversarial objective with respect to a domain discriminator to encourage domain confusion. Representation discriminators include feature discriminator~\cite{chen2017no,tzeng2017adversarial}, output discriminator~\cite{tzeng2017adversarial,tsai2018learning}, conditional discriminator~\cite{long2018conditional}, joint discriminator~\cite{cicek2019unsupervised}, prototypical discriminator~\cite{hu2020panda}, and gradient reversal layers~\cite{ganin2016domain}.
Adversarial generative models combine the domain discriminative model with a generative component based on GAN~\cite{goodfellow2014generative} and its invariants, such as coupled generative adversarial networks (CoGAN)~\cite{liu2016coupled}, SimGAN~\cite{shrivastava2017learning}, and CycleGAN~\cite{zhu2017unpaired,hoffman2018CyCADA,yue2019domain}.
Self-supervision based methods incorporate auxiliary self-supervised learning tasks into the original task network to bring the source and target domains closer. The commonly employed self-supervision visual tasks include reconstruction~\cite{ghifary2015domain,ghifary2016deep,chen2020fido}, image rotation prediction~\cite{sun2019unsupervised,xu2019self}, and jigsaw prediction~\cite{carlucci2019domain}.

\begin{figure*}[!t]
    \centering
    \includegraphics[width=0.88\linewidth]{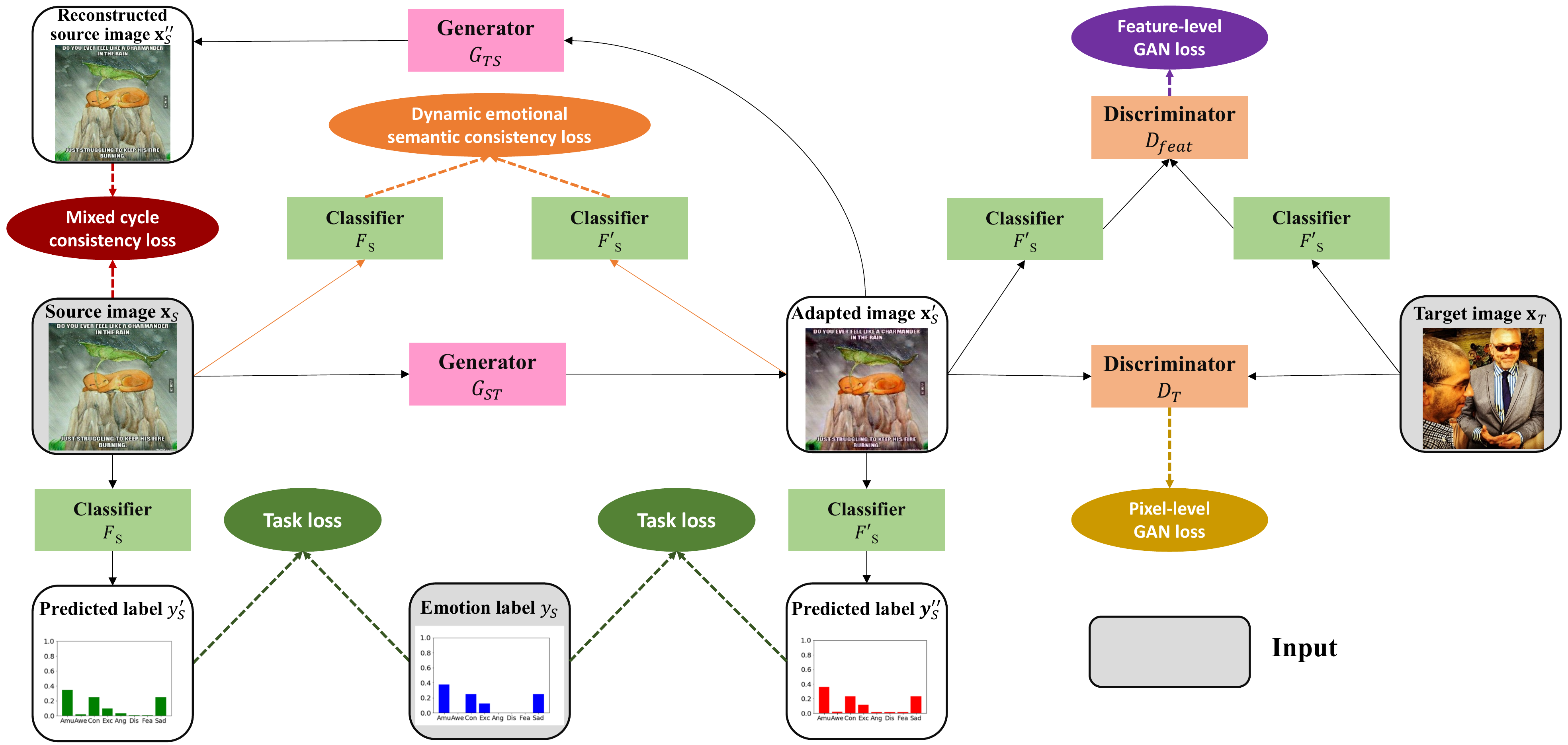} 
    \caption{Framework of the proposed CycleEmotionGAN++ model for visual emotion adaptation from one labeled source domain to another unlabeled target domain. The black solid lines with arrows indicate the operations in the training stage. The dashed lines with arrows correspond to different losses. For clarity the target cycle is omitted.}
    \label{fig:architecture}
\end{figure*}

All these methods focus on objective tasks (\textit{i.e.} with objective labels), such as digit recognition, gaze estimation, object classification, and scene segmentation. \citeauthor{zhao2018emotiongan}~\cite{zhao2018emotiongan} adapted a subjective variable, image emotion, to learn discrete distributions. Later, \citeauthor{zhao2019cycleemotiongan}~\cite{zhao2019cycleemotiongan} studied the UDA problem in image emotion classification. In this paper, we study the UDA problem in both image emotion classification and emotion distribution learning tasks. The comparison between the proposed CycleEmotionGAN++ and existing UDA methods is summarized in Table~\ref{tab:compare}, from which we can see the advantages of CycleEmotionGAN++ relative to other approaches.

\noindent\textbf{Image Style Transfer.} Image style transfer that aims to transfer visual appearance between images is closely related to domain adaptation and has achieved remarkable success recently~\cite{reinhard2001color,wei2004color,hwang2014color,rabin2014adaptive,zhu2017unpaired,yan2016automatic,liu2018emotional,liu2018texture}. \citeauthor{reinhard2001color}~\cite{reinhard2001color} proposed a method by using the $l \alpha \beta$ space to minimize correlation between channels to simplify the transfer process. 
\citeauthor{hwang2014color}~\cite{hwang2014color} proposed a scattered point interpolation scheme using moving least squares to deal with mis-alignments. 
\citeauthor{rabin2014adaptive}~\cite{rabin2014adaptive} proposed image color transfer method based on the relaxed discrete optimal transport techniques.
\citeauthor{zhu2017unpaired}~\cite{zhu2017unpaired} proposed CycleGAN by adding a constraint generator and corresponding loss to assure transfer learning process.
\citeauthor{yan2016automatic}~\cite{yan2016automatic} introduced an image descriptor to achieve semantics-aware photo enhancement (SAPE). 
Some methods specifically designed for emotion color transfer have emerged.
\citeauthor{liu2018emotional}~\cite{liu2018emotional} investigated emotional image color transfer (EICT)
in a network with four modules to make the enhancement results meet user’s emotions.
\citeauthor{liu2018texture}~\cite{liu2018texture} studied texture-aware emotional color
transfer (TAECT) to adjust an image to a reference one by considering the texture information. These image style transfer methods might not perform well for domain adaptation since they do not explicitly align the distributions between different domains.

\section{The Proposed CycleEmotionGAN++ Model}

In this paper, we focus on unsupervised domain adaptation (UDA) for visual emotion analysis (VEA) from one source domain with emotion labels to another target domain without any labels. Suppose the source images and corresponding emotion labels drawn from the source domain distribution $P_{S}(\textbf{x},\textbf{y})$ are $\textbf{x}_{S}$ and $\textbf{y}_{S}$ respectively, and target images drawn from the target domain distribution $P_{T}(\textbf{x})$ are $\textbf{x}_{T}$. Our objective is to train a model that can map an image from the target domain to $L$ (8 in our setting) classes of emotion categories.
%We train two generators $G_{ST}$ and $G_{TS}$, three discriminators $D_{T}$, $D_{S}$ and $D_{feat}$, and two classifiers $F_{S}$ and $F'_{S}$. 

The framework of the proposed CycleEmotionGAN++ model is shown in Fig.~\ref{fig:architecture}. The main idea is to train a mapping network $G_{ST}:\textbf{x}_{S}\rightarrow\textbf{x}_{T}$ which is used to generate adapted images $\textbf{x}'_{S}$ from source images, with the requirement that the adapted images $\textbf{x}'_{S}$ are indistinguishable from the target images $\textbf{x}_{T}$ by the discriminator $D_{T}$. Because the generator mapping $G_{ST}$ is under-constrained and unstable~\cite{zhu2017unpaired}, we impose some constraints. That is, an inverse mapping $G_{TS}$ is employed to reconstruct the source images from the adapted images. A cycle-consistency loss is used to enforce that the reconstructed images $\textbf{x}''_{S}$ and the source images $\textbf{x}_{S}$ are as close as possible. And in order to overcome the drawbacks of a traditional cycle-consistency loss, multi-scale structural similarity is added to the loss. There is a similar cycle from the target to the source. In order to make the adapted images and target images similar, we should ensure that they are similar not only on a pixel-level, but also on a feature-level. So we train another discriminator $D_{feat}$ to perform feature-level alignment. To preserve the emotion labels of the source images, we propose dynamic emotional semantic consistency (DESC) loss with two different classifiers to penalize large semantic differences between the adapted and source images. In this way, the CycleEmotionGAN++ model can adapt the source domain images to be indistinguishable from the target domain, while preserving the annotation information. Finally, we train the task classifier $F'_{S}$ on the adapted dataset $\{\textbf{x}'_{S}$, $\textbf{y}_{S}\}$ by considering that the adapted images $\textbf{x}'_{S}$ and target images $\textbf{x}_{T}$ are from the same distribution.

\subsection{Mixed CycleGAN Loss}
CycleGAN~\cite{zhu2017unpaired} aims to learn two mappings $G_{ST}:\textbf{x}_{S}\rightarrow\textbf{x}_{T}$ and $G_{TS}:\textbf{x}_{T}\rightarrow\textbf{x}_{S}$ between two domains S and T given training samples $\textbf{x}_{S}$ and $\textbf{x}_{T}$. Meanwhile, two discriminators $D_{T}$ and $D_{S}$ are trained, where $D_{T}$ tries to maximize the probability of correctly classifying target images $\textbf{x}_{T}$ and adapted images $\textbf{x}'_{S}$, while the generator $G_{ST}$ tries to generate images to fool $D_{T}$. $D_{S}$ and $G_{TS}$ perform similar operations. As in~\cite{zhu2017unpaired}, the CycleGAN loss contains two terms. One is the adversarial loss~\cite{goodfellow2014generative} that matches the distribution of generated images to the data distribution in the target domain:
\begin{equation}
    \begin{split}
    L_{GAN}&(G_{ST},D_{T},\textbf{x}_{S},\textbf{x}_{T}) = E_{\textbf{x}_{T} \sim P_{T}} [\log D_{T}(\textbf{x}_{T})] \\
    &+ E_{\textbf{x}_{S} \sim P_{S}}[\log(1-D_{T}(G_{ST}(\textbf{x}_{S})))],
    \end{split} 
\label{equ:1}
\end{equation}
\begin{equation}
    \begin{split}
    L_{GAN}&(G_{TS},D_{S},\textbf{x}_{T},\textbf{x}_{S}) = E_{\textbf{x}_{S} \sim P_{S}} [\log D_{S}(\textbf{x}_{S})] \\
    &+ E_{\textbf{x}_{T} \sim P_{T}}[\log(1-D_{S}(G_{TS}(\textbf{x}_{T})))].
    \end{split} 
\label{equ:2}
\end{equation}
The other is a cycle-consistency loss that ensures the learned mappings $G_{ST}$ and $G_{TS}$ are cycle-consistent, preventing them from contradicting each other so that the reconstructed image is close to the original image, which means $G_{TS}(G_{ST}(\textbf{x}_{S})) \approx \textbf{x}_{S}$ and $G_{ST}(G_{TS}(\textbf{x}_{T})) \approx \textbf{x}_{T}$. The difference is penalized by using L1 norm and according to~\cite{zhu2017unpaired}, the cycle-consistency loss is defined as:
\begin{equation}
    \begin{split}
    L_{cyc}&(G_{ST},  G_{TS}) = E_{\textbf{x}_{S} \sim P_{S}} \parallel G_{TS}(G_{ST}(\textbf{x}_{S})) - \textbf{x}_{S} \parallel_{1}\\
    &+E_{\textbf{x}_{T} \sim P_{T}} \parallel G_{ST}(G_{TS}(\textbf{x}_{T})) - \textbf{x}_{T} \parallel_{1}.
    \end{split} 
\label{equ:3}
\end{equation}

According to~\cite{zhao2016loss}, the L1 norm loss can preserve the luminance and color of the images. But it does not perform well in preserving the high-frequency detailed information. Based on a top-down assumption~\cite{zhao2016loss} that the human visual system (HVS) is highly adapted for extracting structural information from the scene, a measure of structural similarity (SSIM) which considers luminance, contrast, and structure information is a good approximation of perceived image quality. By considering HVS, SSIM can get more high-frequency detailed information which preserves the contrast better. In addition, multi-scale structural similarity (MS-SSIM) can overcome the drawback of SSIM in facing different viewing conditions and scales. However, MS-SSIM is not particularly sensitive to uniform biases which can cause changes of brightness or shifts of colors. So we can use MS-SSIM combined with L1 norm as the \textit{mixed cycle-consistency loss} to preserve both the detailed information and brightness of images:
\begin{equation}\small
    \begin{split}
    L_{mixed-cyc}&(G_{ST},  G_{TS}) = \alpha \cdot (L_{MS}(G_{TS}(G_{ST}(\textbf{x}_{S})), \textbf{x}_{S}) \\
    & + L_{MS}(G_{ST}(G_{TS}(\textbf{x}_{T})), \textbf{x}_{T})) + (1-\alpha) \cdot L_{cyc}. \\
    \end{split} 
    \label{equ:7}
\end{equation}
According to~\cite{wang2003multiscale}, MS-SSIM is defined as:
\begin{equation}
    \begin{split}
    L_{MS}(x,y) = [l_{M}(x,y)]^{\alpha_{M}} \cdot \prod_{j=1}^{M} [c_{j}(x,y)]^{\beta_{j}} [s_{j}(x,y)]^{\gamma_{j}},\\
    \end{split} 
    \label{equ:8}
\end{equation}
where the exponents $\alpha_{M}$, $\beta_{j}$ and $\gamma_{j}$ are used to adjust the relative importance of different components and M is the scale number~\cite{wang2003multiscale}. Several parameters in this equation are preserved: luminance, contrast, and structure comparison which are defined as:
\begin{equation}
    \begin{split}
    l(x,y) = \frac{2\mu_{x}\mu_{y}+C_{1}}{\mu^{2}_{x}+\mu^{2}_{y}+C_{1}},\\
    \end{split} 
    \label{equ:4}
\end{equation}
\begin{equation}
    \begin{split}
    c(x,y) = \frac{2\sigma_{x}\sigma_{y}+C_{2}}{\sigma^{2}_{x}+\sigma^{2}_{y}+C_{2}},\\
    \end{split} 
    \label{equ:5}
\end{equation}
\begin{equation}
    \begin{split}
    s(x,y) = \frac{\sigma_{xy}+C_{3}}{\sigma_{x}\sigma_{y}+C_{3}}, \\
    \end{split} 
    \label{equ:6}
\end{equation}
where $\mu_{x}$ and $\mu_{y}$ are mean of x and y, $\sigma^{2}_{x}$ and $\sigma^{2}_{y}$ are variance of x and y, $\sigma_{xy}$ is the covariance of x and y, and $C_{1}$, $C_{2}$ and $C_{3}$ are small constants given by $C_{1}=(K_{1}L)^{2}$, $C_{2}=(K_{2}L)^{2}$ and $C_{3}=C_{2}/2$ respectively. $L$ is the dynamic range of the pixel values, and $K_{1} \ll 1$ and $K_{2} \ll 1$ are two scalar constants. 

Therefore, the objective of the mixed CycleGAN loss is:
\begin{equation}
    \begin{split}
    L_{mCycleGAN} & = L_{GAN}(G_{ST},D_{T},\textbf{x}_{S},\textbf{x}_{T})\\
    &+ L_{GAN}(G_{TS},D_{S},\textbf{x}_{T},\textbf{x}_{S})\\
    &+ \beta L_{mixed-cyc}(G_{ST}, G_{TS}).
    \end{split} 
    \label{equ:10}
\end{equation}
where $\beta$ controls the relative importance of the GAN loss with respect to the mixed cycle-consistency loss.

\begin{figure}[t]
    \begin{center}
    \subfigure[Mikels' wheel]{
    \label{fig:MikelsWheel}
    \includegraphics[width=0.4\linewidth]{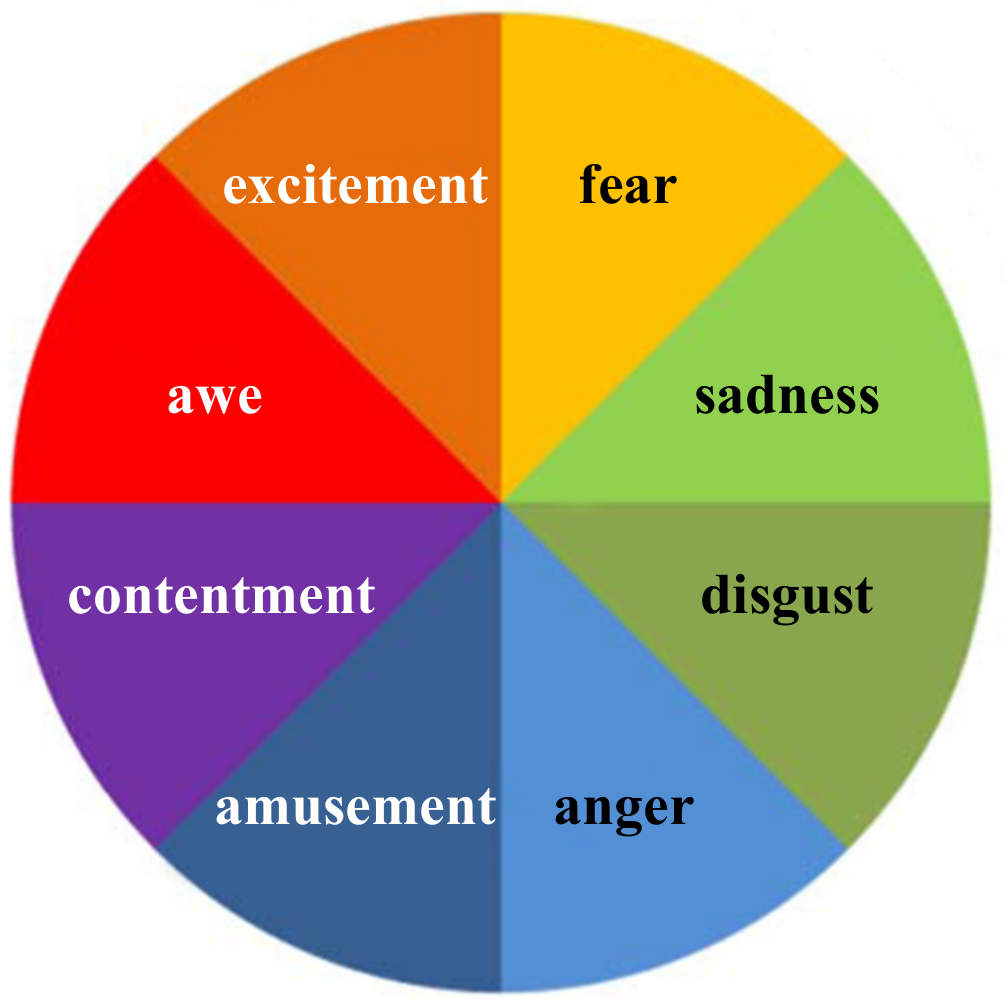}}
    \subfigure[Mikels' emotion distance] {
    \label{fig:MikelsWheelExample}
    \includegraphics[width=0.41\linewidth]{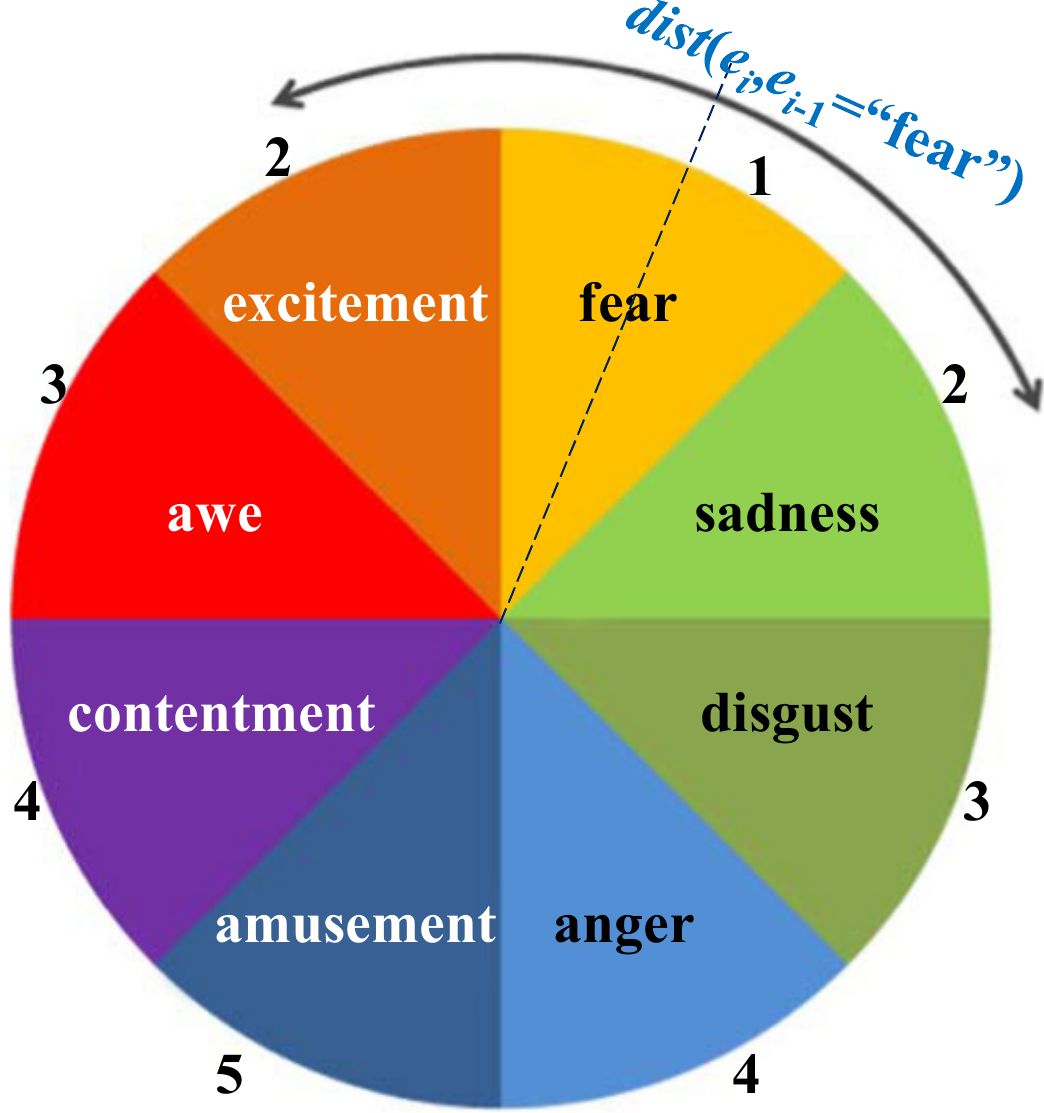}}
    \caption{Mikels' emotion wheel and an example of Mikels’ emotion distances for the emotion category \emph{fear}~\cite{zhao2016predicting}. }
    \label{fig:MikelsWheels}
    \end{center}
\end{figure}

\subsection{Dynamic Emotional Semantic Consistency Loss}
The classifier $F'_{S}$ is trained based on the adapted images and the emotion labels of corresponding source images with the assumption that the emotion labels do not change during the adaptation process. However, this assumption is not always valid. And in order to preserve the emotion labels of the source images, we add a dynamic emotional semantic consistency (DESC) loss. That is, we try to enforce the predicted emotions of the source images $\textbf{x}_{S}$ and adapted images $\textbf{x}'_{S}$ to be as close as possible. And since we have already known that source images and adapted images have different styles, we use two different classifiers to compute the loss. For source images, we use $F_{S}$ which is trained on source domain; for adapted images, we use $F'_{S}$ which is trained on adapted domain. The DESC loss is defined as: 
\begin{equation}
    \begin{split}
    L_{DESC}(G_{ST}) = E_{\textbf{x}_{S} \sim P_{S}}d(F_{S}(\textbf{x}_{S}),F'_{S}(G_{ST}(\textbf{x}_{S}))),\\
    \end{split} 
    \label{equ:11}
\end{equation}
\begin{equation}
    \begin{split}
    L_{DESC}(G_{TS}) = E_{\textbf{x}_{T} \sim P_{T}}d(F'_{S}(\textbf{x}_{T}),F_{S}(G_{TS}(\textbf{x}_{T}))), \\
    \end{split} 
    \label{equ:12}
\end{equation}
where $d(\cdot,\cdot)$ is a function that measures the distance between two emotion labels. In this paper, we define $d$ in two ways. The first one is using symmetric Kullback–Leibler divergence (SKL) to measure the difference of two distributions $\textbf{p}$ and $\textbf{q}$: 
\begin{equation}
    \begin{split}
    SKL(\textbf{p} \parallel \textbf{q}) = KL(\textbf{p} \parallel \textbf{q}) + KL(\textbf{q} \parallel \textbf{p}),\\
    \end{split} 
    \label{equ:13}
\end{equation}
\begin{equation}
    \begin{split}
    KL(\textbf{p} \parallel \textbf{q}) = \sum_{l=1}^L (\textbf{p}_{l} \ln \textbf{p}_{l} - \textbf{p}_{l} \ln \textbf{q}_{l}).\\
    \end{split} 
    \label{equ:14}
\end{equation}
Second, we employ Mikels’ Wheel~\cite{zhao2016predicting} to calculate the distance between emotions. As one of our overall goals is dominant emotion classification, we select the emotion category with the largest probability as our emotion label. Pairwise emotion distance is defined as 1+“the number of steps required to reach one emotion from another”, as shown in Fig.~\ref{fig:MikelsWheels}. Pairwise emotion similarity is defined as the reciprocal of pairwise emotion distance. $d(\cdot,\cdot)$ equals 1-pairwise emotion similarity. From the definitions of these two methods, for the emotion distribution learning task, we can only use the first method as the DESC loss, and for dominant emotion classification, we can use both methods. We name the models using these two methods as CycleEmotionGAN++-SKL and CycleEmotionGAN++-Mikels respectively.

\subsection{Feature-Level Alignment Loss}
Since we want the adapted images $\textbf{x}'_{S}$ and target images $\textbf{x}_{T}$ to be similar, we should ensure that they are similar not only on a pixel-level, but also on a feature-level. Our model trains a discriminator $D_{feat}$ which tries to maximize the probability of correctly classifying adapted images $\textbf{x}'_{S}$ and target images $\textbf{x}_{T}$. The feature-level information we leverage is the output of the last layer in $F'_{S}$. So the information is a $L$-dimension vector. We suppose that the adapted images drawn from the distribution of $F'_{S}(G_{ST}(\textbf{x}_{S}))$ are $\textbf{x}'''_{S}$ and rename the distribution as $P'_{S}$
\begin{equation}
    \begin{split}
    L_{GAN}&(F'_{S},D_{feat},\textbf{x}_{T},F'_{S}(G_{ST}(\textbf{x}_{S}))) = \\
    & E_{\textbf{x}'''_{S} \sim P'_{S}}[\log D_{feat}(F'_{S}(G_{ST}(\textbf{x}_{S})))] \\
    & + E_{\textbf{x}_{T} \sim P_{T}}[\log(1-D_{feat}(F'_{S}(\textbf{x}_{T})))].
    \end{split} 
    \label{equ:19}
\end{equation}
%Based on the model proposed in~\cite{tzeng2017adversarial}, we train $D_{feat}$ after finished training the CycleGAN and two classifiers. We generate the adapted images $\textbf{x}'_{S}$ and initialize the classifier by using $F'_{S}$.

\begin{algorithm}[!t]
{\small
\KwIn{Sets of source images $\textbf{x}_{s}$ $\in$ $\textbf{x}_{S}$ with emotion labels $\textbf{y}_{s}$ $\in$ $\textbf{y}_{S}$, target images $\textbf{x}_{t}$ $\in$ $\textbf{x}_{T}$, the maximum number of steps of the first and second parts $T_{1}$, $T_{2}$ respectively, a threshold \textbf{$thres$}.}

\KwOut{Predicted emotion label distributions of target domain image $\textbf{x}_{T}$.}

\For{$i\leftarrow 1\ \textbf{to}\ T_{1}$}{
    Sample a mini-batch of source images $\textbf{x}_{s}$ and target images $\textbf{x}_{t}$.

    %\vspace{6pt}

    \tcc{Updating $\theta_{ST}$ and $\theta_{TS}$ when fixing $\phi_{S}$, $\phi_{T}$, $\delta_{S}$ and $\delta'_{S}$}

    Update $\theta_{ST}$ and $\theta_{TS}$ by taking an SGD step on mini-batch loss $L_{mCycleGAN}$ plus $L_{DESC}(G_{ST})$ and $L_{DESC}(G_{TS})$ in Eq.~(\ref{equ:10}), Eq.~(\ref{equ:11}) and Eq.~(\ref{equ:12}). 

    %\vspace{6pt}

    \tcc{Updating $\phi_{S}$, $\phi_{T}$ when fixing $\theta_{ST}$, $\theta_{TS}$, $\delta_{S}$ and $\delta'_{S}$}

    Compute $G_{ST}(\textbf{x}_{s}, \theta_{ST})$ with current $\theta_{ST}$.

    Compute $G_{TS}(\textbf{x}_{t}, \theta_{TS})$ with current $\theta_{TS}$.

    Update $\phi_{T}$ and $\phi_{S}$ by taking an SGD step on mini-batch
loss $L_{GAN}$ in Eq.~(\ref{equ:1}), Eq.~(\ref{equ:2}). 

    %\vspace{6pt}

    \tcc{Updating $\delta_{S}$ and $\delta'_{S}$ when fixing $\theta_{ST}$, $\theta_{TS}$, $\phi_{S}$ and $\phi_{T}$}

    Compute $G_{ST}(\textbf{x}_{s}, \theta_{ST})$ with current $\theta_{ST}$.
    
    Update $\delta_{S}$, $\delta'_{S}$ by taking an SGD step on mini-batch loss $L_{task}(F_{S})$ and $L_{task}(F'_{S})$ in Eq.~(\ref{equ:15}) / Eq.~(\ref{equ:17}) and Eq.~(\ref{equ:16}) / Eq.~(\ref{equ:18}).
}

Compute $G_{ST}(\textbf{x}_{s}, \theta_{ST})$ with current $\theta_{ST}$.

\For{$j\leftarrow 1\ \textbf{to}\ T_{2}$}{
    Update $\phi$ by taking an SGD step on mini-batch loss $L_{GAN}$ and in Eq.~(\ref{equ:19}).
    
    \If{$Accuracy(D_{feat}) > thres$}{
        Update $\delta'_{S}$ by taking an SGD step on mini-batch loss $L_{task}(F'_{S})$ in Eq.~(\ref{equ:16}) / Eq.~(\ref{equ:18}).}
}
\Return $F'_{S}(\textbf{x}_{t}, \delta'_{S})$;}

\small\caption{Adversarial training procedure of our CycleEmotionGAN++ model}
\label{alg:model}
\end{algorithm}

\subsection{Task Classification Loss}
Under the assumption that the emotion labels of the adapted images do not change during the adaptation process, we can train a transferable task classifier $F'_{S}$ based on the adapted images $\textbf{x}'_{S}$ and corresponding source emotion labels $\textbf{y}_{S}$. Besides $F'_{S}(\textbf{x}'_{s}) \rightarrow \textbf{y}''_{s}$, which assigns emotion $\textbf{y}''_{S}$ to the adapted image $\textbf{x}'_{s}$, the proposed CycleEmotionGAN++ is augmented with another classifier $F_{S}(\textbf{x}_{s}) \rightarrow \textbf{y}'_{s}$ assigning emotion $\textbf{y}'_{S}$ to the source image $\textbf{x}_{s}$ for dynamic emotional semantic consistency.
%Generally, another two separate task models are learned based on the adapted source images $\textbf{x}'_{S}$ and the corresponding emotion labels $\textbf{y}_{S}$ under the assumption that the emotion labels of the adapted images do not change during the adaptation process for classifier $F'_{S}$ and the source images $\textbf{x}_{S}$ and the corresponding emotion labels $\textbf{y}_{S}$ for classifier $F_{S}$. Contrary to this, the proposed CycleEmotionGAN++ is augmented with two classifiers: $F'_{S}(\textbf{x}'_{s}) \rightarrow \textbf{y}''_{s}$, which assigns emotion $\textbf{y}''_{S}$ to the adapted image $\textbf{x}'_{s}$ and $F_{S}(\textbf{x}_{s}) \rightarrow \textbf{y}'_{s}$, which assigns emotion $\textbf{y}'_{S}$ to the adapted image $\textbf{x}_{s}$. 
For emotion distribution learning, we use the KL-Divergence as the task loss:
\begin{equation}
    \begin{split}
    L_{task}(F_{S}) = E_{(\textbf{x}_{S},\textbf{y}_{S}) \sim P_{S}} KL(\textbf{y}_{S} \parallel F_{S}(\textbf{x}_{S})),\\
    \end{split} 
    \label{equ:15}
\end{equation}
\begin{equation}
    \begin{split}
    L_{task}(F'_{S}) = E_{(\textbf{x}_{S},\textbf{y}_{S}) \sim P_{S}} KL(\textbf{y}_{S} \parallel F'_{S}(G_{ST}(\textbf{x}_{S}))). \\
    \end{split} 
    \label{equ:16}
\end{equation}
For dominant emotion classification, following~\cite{you2016building}, the two classifiers $F_{S}$ and $F'_{S}$ are optimized by minimizing the standard cross-entropy loss:
\begin{equation}
    \small
    \begin{split}
L_{task}(F_{S})={E}_{(\textbf{x}_S,\textbf{y}_S)\sim P_S}\sum_{l=1}^{L}{\mathds{1}}_{[l=\textbf{y}_S]}\log(\sigma(F_{S}^{(l)}(\textbf{x}_S))),
    \end{split} 
    \label{equ:17}
\end{equation}
\begin{equation}
    \small
    \begin{split}
L_{task}(F'_{S})={E}_{(\textbf{x}_S,\textbf{y}_S)\sim P_S}\sum_{l=1}^{L}{\mathds{1}}_{[l=\textbf{y}_S]}\log(\sigma(F_{S}^{'(l)}(G_{ST}(\textbf{x}_S)))).
    \end{split} 
    \label{equ:18}
\end{equation}
where $\sigma$ is the softmax function, and $\mathds{1}$ is an indicator function.

\begin{table*}  
\centering\footnotesize
\caption{Classification accuracy ($\%$) comparison between CycleEmotionGAN++ and state-of-the-art approaches when adapting from ArtPhoto to FI. The best accuracy of each emotion category and the average accuracy are emphasized in bold.}
\begin{center}  
\begin{tabular}{c|cccccccc|c }  
\toprule
Method & Amu & Ang & Awe & Con & Dis & Exc & Fea & Sad & Avg\\
\hline  
Source-only & 47.63 & 2.83 & 25.86 & 6.33 & 5.57 & 8.67 & 16.50 & 51.15 & 23.86\\
CycleGAN~\citep{zhu2017unpaired} & 28.19 & 20.24 & 13.58 & 29.87 & 30.90 & 21.59 & 25.00 & 32.86 & 25.99\\
SAPE~\cite{yan2016automatic} & 30.89 & 17.32 & 21.26 & 25.99 & 9.82 & 33.68 & 17.00 & 31.45 & 26.04 \\
EICT~\cite{liu2018emotional} & 30.08 & 8.66 & 16.28 & 33.51 & 2.45 & 24.56 & 15.00 & 24.73 & 24.00\\
TAECT~\cite{liu2018texture} & 36.69 & 7.87 & 22.92 & 21.68 & 14.11 & \textbf{34.39} & 19.00 & 20.49 & 25.09\\
ADDA~\citep{tzeng2017adversarial} & 39.51 & 1.62 & 32.90 & 8.13 & 4.33 & 7.96 & 7.50 & \textbf{79.22} & 26.33\\
SimGAN~\citep{shrivastava2017learning} & 23.58 & 3.94 & 21.54 & \textbf{34.95} & 30.06 & 29.82 & 10.00 & 27.91 & 26.13\\
CyCADA~\citep{hoffman2018CyCADA} & 33.74 & 18.90 & 22.83 & 44.27 & 23.31 & 10.18 & 14.00 & 34.63 & 29.62\\
CycleEmotionGAN-Mikels (ours)& 42.90 & 4.45 & \textbf{41.41} & 5.10 & 6.81 & 15.93 & 3.00 & 74.25 & 28.00 \\
CycleEmotionGAN-SKL (ours)& 55.14 & 12.96 & 36.50 & 4.73 & 6.19 & 4.96 & 1.00 & 63.40 & 27.50 \\
\textbf{CycleEmotionGAN++-Mikels (ours)} & \textbf{66.26} & 16.54 & 21.54 & 27.78 & 3.68 & 14.39 & 6.00 & 40.28 & 31.74\\
\textbf{CycleEmotionGAN++-SKL (ours)} & 44.86 & \textbf{40.49} & 18.33 & 32.99 & \textbf{30.96} & 17.88 & \textbf{50.00} & 27.53 & \textbf{32.01}\\
\hline 
Oracle (train on target) & 77.24 & 44.88 & 72.03 & 65.59 & 60.12 & 61.40 & 48.00& 65.37 & 66.11\\
\bottomrule 
\end{tabular}  
\end{center} 
\label{tab:art-fi}   
\end{table*}

\begin{table*}  
\centering\footnotesize
\caption{Classification accuracy ($\%$) comparison between CycleEmotionGAN++ and state-of-the-art approaches when adapting from FI to ArtPhoto.}  
\begin{center}  
\begin{tabular}{c|cccccccc|c } 
\toprule
Method & Amu & Ang & Awe & Con & Dis & Exc & Fea & Sad & Avg\\
\hline  
Source-only & \textbf{30.00} & 28.57 & \textbf{55.00} & 35.71 & 14.29 & 20.00 & 18.18 & 30.30 & 29.11\\
CycleGAN~\citep{zhu2017unpaired} & 15.00 & 28.57 & 20.00 & 7.14 & 21.43 & \textbf{60.00} & 40.90 & 42.42 & 31.65\\
SAPE~\cite{yan2016automatic} & \textbf{30.00} & 7.14 & 45.00 & 35.71 & 7.14 & \textbf{60.00} & 30.43 & 54.55 & 33.54 \\
EICT~\cite{liu2018emotional} & 25.00 & 28.57 & 30.00 & 21.43 & 14.29 & 50.00 & 26.09 & 42.42 & 31.64\\
TAECT~\cite{liu2018texture} & 20.00 & 14.29 & 40.00 & 21.43 & 28.57 & 55.00 & 30.43 & 30.30 & 31.01\\
ADDA~\citep{tzeng2017adversarial} & 25.00 & \textbf{42.86} & 40.00 & 7.14 & 14.29 & 55.00 & 40.90 & 30.30 & 32.91\\
SimGAN~\citep{shrivastava2017learning} & 15.00 & 14.29 & 45.00 & 28.57 & 7.14 & 25.00 & 31.82 & \textbf{63.64} & 32.91\\
CyCADA~\citep{hoffman2018CyCADA} & 20.00 & 21.43 & 45.00 & 0.00 & 35.71 & 40.00 & 59.09 & 57.58 & 38.61\\
CycleEmotionGAN-Mikels (ours)& \textbf{30.00} & \textbf{42.86} & 45.00 & 21.43 & 21.43 & 50.00 & 55.00 & 33.33 & 37.97 \\
CycleEmotionGAN-SKL (ours)& 25.00 & 28.57 & 35.00 & \textbf{42.86} & 0.00 & 55.00 & 54.55 & 42.42 & 37.34 \\
\textbf{CycleEmotionGAN++-Mikels (ours)} & 25.00 & 28.57 & 30.00 & \textbf{42.86} & \textbf{42.86} & 55.00 & 72.73 & 27.27 & 39.87\\
\textbf{CycleEmotionGAN++-SKL (ours)} & \textbf{30.00} & 36.00 & 40.00 & 21.43 & 14.29 & 40.00 & \textbf{77.27} & 45.45 & \textbf{40.51}\\
\hline 
Oracle (train on target) & 55.00 & 35.71 & 30.00 & 14.29 & 42.86 & 55.00 & 59.09 & 45.45 & 43.67\\
\bottomrule
\end{tabular}  
\end{center} 
\label{tab:fi-art}
\end{table*}

\subsection{CycleEmotionGAN++ Learning}

Our model objective loss combines the CycleGAN loss, DESC loss and the feature-level alignment loss:
\begin{equation}
\small
    \begin{split}
    & L_{Model} = L_{mCycleGAN}+ \gamma L_{DESC}(G_{ST})\\
    &+ \gamma L_{DESC}(G_{TS})+ L_{GAN}(F'_{S},D_{feat},\textbf{x}_{T},F'_{S}(G_{ST}(\textbf{x}_{S}))).
    \end{split} 
    \label{equ:20}
\end{equation}
where $\gamma$ controls the relative importance of the DESC loss to the overall loss.

%\subsection{Model learning}
In our implementation, the generators $G_{ST}$ and $G_{TS}$ are CNNs with residual connections that maintain the resolution of the original image as illustrated in Fig.~\ref{fig:architecture}. The discriminators $D_{T}$, $D_{S}$, $D_{feat}$ and the classifiers $F_{S}$, $F'_{S}$ are also CNNs. The optimization of the our model is divided into two parts. In the first part, we optimize $G_{ST}$, $G_{TS}$, $D_{T}$, $D_{S}$, $F_{S}$, $F'_{S}$. Those networks are optimized by alternating between three stochastic gradient descent (SGD) steps. During the first step, we fix $D_{T}$, $D_{S}$, $F_{S}$, $F'_{S}$ and update $G_{ST}$, $G_{TS}$. During the second step, we update $D_{T}$, $D_{S}$, while keeping $G_{ST}$, $G_{TS}$, $F_{S}$ and $F'_{S}$ fixed. During the third step, we update $F_{S}$ and $F'_{S}$, while keeping $G_{ST}$, $G_{TS}$, $D_{T}$, $D_{S}$ fixed. After the first part, we use $G_{ST}$ to generate the adapted domain $\textbf{x}'_{S}$ with $G_{ST}(\textbf{x}_{S})$. In the second part, we also optimize $D_{feat}$ and $F'_{S}$ by using SGD steps. $F'_{S}$ is fixed when $D_{feat}$'s accuracy is lower than 0.8. The detailed training procedure is shown in Algorithm~\ref{alg:model}, where $\theta_{ST}$, $\theta_{TS}$, $\phi_{S}$, $\phi_{T}$, $\delta_{S}$, $\delta'_{S}$ and $\phi$ are the parameters of $G_{ST}$, $G_{TS}$, $D_{S}$, $D_{T}$, $F_{S}$, $F'_{S}$ and $D_{feat}$, respectively.

\section{Experiments}
In this section, we introduce the experimental settings, evaluate the performance of the proposed model, and report and analyze the results as compared to state-of-the-art approaches.
%In this section, we first introduce the detailed experimental settings, including the datasets, evaluation metrics, baselines, and implementation details. We then evaluate the performance of the proposed model, and report and analyze the results as compared to state-of-the-art approaches.

\subsection{Datasets}
\textbf{Flickr-LDL}~\cite{yang2017learning} is a subdataset of FlickrCC~\cite{borth2013large} and contains 11,500 images which are labeled by 11 viewers based on Mikels’ eight emotion categories. \textbf{Twitter-LDL}~\cite{yang2017learning} contains 10,045 images obtained by searching from Twitter using emotions keywords. The images are labeled by 8 viewers also based on Mikels’ emotion categories. The original labels of each image in Flickr-LDL and Twitter-LDL datasets are the number of votes on each emotion category. To obtain the emotion distribution labels, we divide the vote of each category by the total number of voters.

\textbf{ArtPhoto}~\cite{machajdik2010affective} contains 806 artistic photographs organized by Mikels’ emotion categories. The photographers took the photos, uploaded them to the website and determined which one of the eight emotion categories each photo belonged to. The artists try to evoke a certain emotion in the viewers through the photos with conscious manipulation of the emotional objects, lighting, colors, \emph{etc}. The Flickr and Instagram (\textbf{FI}) dataset~\cite{you2016building} contains
% is made among 3 million 
images from the Flickr and Instagram websites which are labeled into one of Mikels' emotion categories by a group of 225 Amazon Mechanical Turk (AMT) workers. 23,308 images which received at least three agreements among workers are included in the FI dataset.

\begin{table*}  
\centering\footnotesize
\caption{Comparison of CycleEmotionGAN++ with state of-the-art methods when adapting from the source domain Twitter-LDL to the target domain Flickr-LDL. The best method trained on the source domain is emphasized in bold.}
\begin{center} 
\begin{tabular}{c|cccccc}  
\toprule
Method & $SSD\downarrow$ & $KL\downarrow$ & $BC\uparrow$ & $Canbe\downarrow$ & $Cheb\downarrow$ & $Cos\uparrow$ \\
\hline  
Source-only & 0.1864 & 0.6845 & 0.7974 & 5.8347 & 0.2964 & 0.7846\\
CycleGAN~\citep{zhu2017unpaired} & 0.1781 & 0.6441 & 0.8000 & 5.8173 & 0.2905 & 0.7904\\
SAPE~\cite{yan2016automatic} & 0.1768 & 0.6375 & 0.8045 & 5.8434 & 0.2875 & 0.7935\\
EICT~\cite{liu2018emotional} & 0.1803 & 0.6614 & 0.8129 & 5.9373 & 0.2904 & 0.8102\\
TAECT~\cite{liu2018texture} & 0.1777 & 0.6589 & 0.8046 & 5.9027 & 0.2843 & 0.8001\\
ADDA~\citep{tzeng2017adversarial} & 0.1876 & 0.6489 & \textbf{0.8304} & 6.0150 & 0.2830 & 0.8026\\
SimGAN~\citep{shrivastava2017learning}& 0.1787 & 0.6505 & 0.8073 & 5.8621 & 0.2891 & 0.7904\\
CyCADA~\citep{hoffman2018CyCADA} & \textbf{0.1589} & 0.5914 & 0.8130 & 5.7576 & 0.2757 & 0.8126\\
CycleEmotionGAN-SKL (Ours) & 0.1672 & 0.6123 & 0.8156 & 5.8374 & 0.2775 & 0.8089\\
\textbf{CycleEmotionGAN++-SKL (ours)} & \textbf{0.1589} & \textbf{0.5772} & 0.8214 & \textbf{5.7542} & \textbf{0.2718} & \textbf{0.8167}\\
\hline 
Oracle (train on target) & 0.1419 & 0.5306 & 0.8248 & 5.5078 & 0.2597 & 0.8335\\
\bottomrule
\end{tabular}  
\end{center} 
\label{tab:twi-fli}  
\end{table*}

\begin{table*}  
\centering\footnotesize
\caption{Comparison of CycleEmotionGAN++ with state-of-art methods when adapting from Flickr-LDL to Twitter-LDL.}
\begin{center}  
\begin{tabular}{c|cccccc}  
\toprule 
Method & $SSD\downarrow$ & $KL\downarrow$ & $BC\uparrow$ & $Canbe\downarrow$ & $Cheb\downarrow$ & $Cos\uparrow$ \\
\hline  
Source-only & 0.1856 & 0.6650 & 0.7791 & 6.0716 & 0.3066 & 0.8004\\
CycleGAN~\citep{zhu2017unpaired} & 0.1712 & 0.6392 & 0.7964 & 6.0234 & 0.2896 & 0.8132\\
SAPE~\cite{yan2016automatic} & 0.1694 & 0.6101 & 0.8086 & 6.0700 & 0.2799 & 0.8263\\
EICT~\cite{liu2018emotional} & 0.1793 & 0.6482 & 0.7819 & 6.0652 & 0.2865 & 0.8097\\
TAECT~\cite{liu2018texture} & 0.1735 & 0.6392 & 0.7916 & 6.0635 & 0.2803 & 0.8139\\
ADDA~\citep{tzeng2017adversarial} & 0.1693 & 0.6306 & 0.7924 & 6.0630 & 0.2883 & 0.8141\\
SimGAN~\citep{shrivastava2017learning}& 0.1751 & 0.6338 & 0.7919 & 6.0560 & 0.2938 & 0.8088\\
CyCADA~\citep{hoffman2018CyCADA} & 0.1493 & 0.5617 & 0.8120 & \textbf{6.0178} & 0.2712 & 0.8375\\
CycleEmotionGAN-SKL(ours)& 0.1541 & 0.5800 & 0.8124 & 6.0512 & 0.2688 & 0.8327\\
\textbf{CycleEmotionGAN++-SKL (ours)} & \textbf{0.1410} & \textbf{0.5412} & \textbf{0.8273} & 6.0336 & \textbf{0.2529} & \textbf{0.8462}\\
\hline 
Oracle (train on target) & 0.1274 & 0.5003 & 0.8439 & 5.8543 & 0.2389 & 0.8629\\
\bottomrule
\end{tabular}  
\end{center} 
\label{tab:fli-twi}  
\end{table*}

\subsection{Evaluation Metrics}
\label{subsec:evaluation}
\subsubsection{Twitter-LDL and Flickr-LDL}
We use different metrics to evaluate the performance of our model: the sum of squared difference (SSD)~\cite{zhao2017approximating}, Kullback-Leibler (KL) divergence\footnote{\url{https://en.wikipedia.org/wiki/Kullback\%E2\%80\%93Leibler\_divergence}}, Bhattacharyya coefficient (BC)\footnote{\url{https://en.wikipedia.org/wiki/Bhattacharyya\_distance}}, Canberra distance (Canbe)\footnote{\url{https://en.wikipedia.org/wiki/Canberra\_distance}}, Chebyshev distance (Cheb)\footnote{\url{https://en.wikipedia.org/wiki/Chebyshev\_distance}}, and Cosine similarity (Cos)\footnote{\url{https://en.wikipedia.org/wiki/Cosine\_similarity}}. For BC and Cos, larger values indicate better results and for SSD, KL, Canbe, and Cheb, smaller values indicate better results. KL is leveraged as the main metric.% for performance evaluation. 
% SSD measures the regression performance ranging from 0 to 1:
% \begin{equation}
%     \begin{split}
%     SSD(u,v)\downarrow = \sum_{(x,y)\in N}[I(u+x,v+y)-P(x,y)]^2\\
%     \end{split} 
%     \label{equ:21}
% \end{equation}
% KL, BC, Canbe, and Cheb measure the distance between two distributions, and are defined as: 
% % $KL \geq 0$, $0 \leq BC \leq 1$, $Canbe \geq 0$, $0 \leq Cheb \leq 1$ respectively.
% \begin{equation}
%     \begin{split}
%     BC(p,q)\uparrow = \sum_{x\in X}\sqrt{p(x)q(x)}\\
%     \end{split} 
%     \label{equ:22}
% \end{equation}
% \begin{equation}
%     \begin{split}
%     Canbe(p,q)\downarrow = \sum_{i=1}^{n} \frac{|p_{i}-q_{i}|}{|p_{i}|+|q_{i}|}\\
%     \end{split} 
%     \label{equ:23}
% \end{equation}
% \begin{equation}
%     \begin{split}
%     Cheb(x,y)\downarrow = \max_{i}(|x_{i}-y_{i}|)\\
%     \end{split} 
%     \label{equ:24}
% \end{equation}
% Cos measures the similarity between two vectors: % ranging from 0 to 1.
% \begin{equation}
%     \begin{split}
%     Cos(\theta)\uparrow &= \frac{A \cdot B}{\parallel A \parallel \parallel B \parallel}\\
%     &= \frac{\sum_{i=1}^{n} A_{i} B_{i}}{\sqrt{\sum_{i=1}^{n}A_{i}^{2}} \sqrt{\sum_{i=1}^{n}B_{i}^2}}
%     \end{split} 
%     \label{equ:25}
% \end{equation}

\subsubsection{ArtPhoto and FI}
Similar to~\cite{you2016building}, we employ the emotion classification accuracy ($Acc$) as the evaluation metric. $Acc$ is defined as the proportion of correct predictions out of total predictions. We compute the accuracy for each category and employ the average accuracy as the main metric. 

\subsection{Baselines}
To the best of our knowledge, CycleEmotionGAN++ is the first
work on unsupervised domain adaptation for both emotion distribution learning and dominant emotion classification. To demonstrate its effectiveness, we compare it with the following baselines.
%To demonstrate its effectiveness, we compare it to the following baselines which from 2nd to 5th are color transfer methods, while from 6th to 8th are domain adaptation methods.
\textbf{(1) Source-only}: a lower bound which trains a model on the source domain and tests it directly on target domain. 
\textbf{(2) Color style transfer methods}: \textit{CycleGAN}~\cite{zhu2017unpaired}: first translating the source images into adapted images via CycleGAN and then training a task classifier on the adapted images; \textit{SAPE}~\cite{yan2016automatic}: first using descriptor to generate semantics-aware photo enhanced style images and then training a task classifier on the enhanced images; \textit{EICT}~\cite{liu2018emotional}: first translating source images' style and then training a task classifier on the adapted images; and \textit{TAECT}~\cite{liu2018texture}: first transferring source images to the color in the database extracted from reference images and then training a task classifier on the adapted images. 
\textbf{(3) UDA methods}:
\textit{ADDA}~\cite{tzeng2017adversarial}: first training the task classifier on the source images and then aligning the feature-level information of the source and target domains; \textit{SimGAN}~\cite{shrivastava2017learning}: first translating the source images into the target style using the generator augmented with a self-regularization loss, and then training on the adapted images with corresponding source labels; and \textit{CyCADA}~\cite{hoffman2018CyCADA}: first translating source images into the target style with cycle-consistency loss and semantic-consistency loss, and then training the task classifier on the adapted images with feature-level alignment. 
\textbf{(4) Oracle}: an upper bound where the classifier is both trained and tested on the target domain.

\begin{table*} 
\centering\footnotesize
\caption{Ablation study on different components of CycleEmotionGAN++ for emotion distribution learning. Baseline denotes pixel-level alignment with cycle-consistency, +DESC denotes adding dynamic emotional semantic consistency loss, +Feat denotes adding feature-level alignment, +msssim denotes adding multi-scale structure similarity.}  
\begin{center} 
\begin{tabular}{c|c|cccccc}   
\toprule
DA setting & Method & $SSD\downarrow$ & $KL\downarrow$ & $BC\uparrow$ & $Canbe\downarrow$ & $Cheb\downarrow$ & $Cos\uparrow$ \\
\hline  
\multirow{5}*{Twitter-LDL $\rightarrow$ Flickr-LDL} & CycleGAN (Baseline) & 0.1781 & 0.6441 & 0.8000 & 5.8173 & 0.2905 & 0.7904\\
& +DESC & 0.1592 & 0.6027 & 0.8184 & 5.8149 & 0.2732 & 0.8156\\
& +DESC+Feat & \textbf{0.1558} & 0.5901 & \textbf{0.8245} & 5.7745 & 0.2713 & 0.8145\\
& +DESC+msssim & 0.1578 & 0.5949 & 0.8196 & 5.8100 & \textbf{0.2707} & 0.8164\\
& +DESC+msssim+Feat & 0.1589 & \textbf{0.5772} & 0.8214 & \textbf{5.7542} & 0.2718 & \textbf{0.8167}\\
\midrule
\multirow{5}*{Flickr-LDL $\rightarrow$ Twitter-LDL} & CycleGAN (Baseline) & 0.1712 & 0.6392 & 0.7964 & \textbf{6.0234} & 0.2896 & 0.8132\\
& +DESC & 0.1478 & 0.5724 & 0.8176 & 6.0701 & 0.2627 & 0.8395 \\
& +DESC+Feat & 0.1477 & 0.5513 & 0.8256 & 6.0516 & 0.2593 & 0.8400\\
& +DESC+msssim & 0.1498 & 0.5642 & 0.8228 & 6.0591 & 0.2641 & 0.8402\\
& +DESC+msssim+Feat & \textbf{0.1410} & \textbf{0.5412} & \textbf{0.8273} & 6.0336 &\textbf{0.2529} & \textbf{0.8462}\\
\bottomrule
\end{tabular}  
\end{center} 
\label{tab:ablation1} 
\end{table*}

\begin{table*}  
\centering\footnotesize
\caption{Ablation study on different components of CycleEmotionGAN++ for dominant emotion classification.}  
\begin{center}  
\begin{tabular}{c|c|cccccccc|c}  
\toprule
DA setting & Method & Amu & Ang & Awe & Con & Dis & Exc & Fea & Sad & Avg\\
\hline  
\multirow{5}*{ArtPhoto $\rightarrow$ FI} & CycleGAN (Baseline) & 28.19 & 20.24 & 13.58 & 29.87 & 30.90 & 21.59 & 25.00 & 32.86 & 25.99\\
& +DESC & 46.95 & 26.62 & 32.48 & 33.69 & \textbf{38.65} & 31.23 & 8.00 & 9.54 & 27.94\\
& +DESC+Feat & 2.64 & 21.26 & \textbf{38.54} & \textbf{37.10} & 12.27 & 23.51 & 11.00 & \textbf{47.35} & 30.19\\
& +DESC+msssim & \textbf{51.95} & 24.70 & 19.97 & 31.38 & 21.67 & \textbf{35.75} & 13.50 & 7.99 & 30.05\\
& +DESC+msssim+Feat & 44.86 & \textbf{40.49} & 18.33 & 32.99 & 30.96 & 17.88 & \textbf{50.00} & 27.53 & \textbf{32.01}\\
\midrule
\multirow{5}*{FI $\rightarrow$ ArtPhoto} & CycleGAN (Baseline) & 15.00 & 28.57 & 20.00 & 7.14 & 21.43 & \textbf{60.00} & 40.90 & 42.42 & 31.65\\
& +DESC & \textbf{30.00} & 7.14 & 35.00 & \textbf{35.71} & \textbf{35.71} & 55.00 & 36.36 & 51.52 & 37.97\\
& +DESC+Feat & 25.00 & 14.29 & \textbf{50.00} & \textbf{35.71} & 7.14 & 40.00 & 50.00 & \textbf{57.58} & 39.24\\
& +DESC+msssim & 25.00 & 28.57 & \textbf{50.00} & 14.29 & 21.43 & 50.00 & 59.09 & 42.42 & 38.61\\
& +DESC+msssim+Feat & \textbf{30.00} & \textbf{36.00} & 40.00 & 21.43 & 14.29 & 40.00 & \textbf{77.27} & 45.45 & \textbf{40.51}\\
\bottomrule
\end{tabular}  
\end{center} 
\label{tab:ablation3} 
\end{table*}

\subsection{Implementation Details}
The generators $G_{ST}$ and $G_{TS}$ use the model in~\cite{zhu2017unpaired} which shows impressive results for neural style transfer and super resolution. $G_{ST}$ and $G_{TS}$ contain two stride-2 convolutions, several residual blocks, and two fractionally-strided convolutions with stride $\frac{1}{2}$. Our model uses 9 blocks for 256$\times$256 training images. For normalization, we choose instance normalization~\cite{johnson2016perceptual}. The discriminators $D_{T}$ and $D_{S}$ are deployed by using 70$\times$70 PatchGAN~\cite{li2016precomputed} which aims to classify whether 70$\times$70 overlapping image patches are real or fake. Such a patch-level discriminator architecture has fewer parameters than a full-image discriminator and can work on arbitrarily-sized images in a fully convolutional fashion with good performance. The classifiers $F_{S}$ and $F'_{S}$ use the ResNet-101~\cite{he2016deep} architecture pretrained on ImageNet. We finetune the model and update the classification loss into Eq.~(\ref{equ:15})~-~(\ref{equ:18}). The discriminator $D_{feat}$ uses the architecture as~\cite{hoffman2018CyCADA} which contains 3 linear layers mapping a $L$-dimension vector to a 2-dimension one. The GAN loss keeps the same as standard ones. As shown in Algorithm~\ref{alg:model}, $D_{feat}$ is trained after acquiring finetuned $F'_{S}$ and $x'_{S}$ generated by $G_{ST}$.

Similar to LSGAN~\cite{mao2017least}, our model trains a GAN loss $L_{GAN}(G_{ST},D_{T},\textbf{x}_{S},\textbf{x}_{T})$ by minimizing $G_{ST}$ in $E_{\textbf{x}_{S} \sim P_{S}}[(D_{T}(G_{ST}(\textbf{x}_{S}))$-$1)^{2}]$ and minimizing
$D_{T}$ in $E_{\textbf{x}_{T} \sim P_{T}}[(D_{T}(\textbf{x}_{T})$-$1)^{2}]$ + $E_{\textbf{x}_{S}\sim P_{S}}[D_{T}(G_{ST}(\textbf{x}_{S}))^{2}]$. It is more stable to train GANs on high-resolution images. $L_{GAN}(G_{TS},D_{S},\textbf{x}_{T},\textbf{x}_{S})$ is also optimized by using LSGAN. We follow~\cite{shrivastava2017learning} to reduce the model oscillation by updating the discriminators using a history of generated images rather than the ones produced by the latest generators. We leverage an image pool that stores the 50 recently created images.

$\alpha$, $\beta$, and $\gamma$ in Eq.~(\ref{equ:7}), Eq.~(\ref{equ:10}), and Eq.~(\ref{equ:20}) are empirically set to 0.5, 10, and 50 respectively. Similar to~\cite{wang2003multiscale}, in order to simplify parameter selection, we set $\alpha_{j}$=$\beta_{j}$=$\gamma_{j}$ for all $j$’s in Eq.~(\ref{equ:8}), and normalize the cross-scale settings such that $\sum_{j=1}^{M}\gamma_{j}=1$, which makes different parameter settings comparable. M is set to 5 and $K_{1}$, $K_{2}$ are set to 0.01, 0.03 respectively. We also set $\beta_{1}$ = $\gamma_{1}$ = 0.0448, $\beta_{2}$ = $\gamma_{2}$ = 0.2856, $\beta_{3}$ = $\gamma_{3}$ = 0.3001, $\beta_{4}$ = $\gamma_{4}$ = 0.2363, and $\alpha_{5}$ = $\beta_{5}$ = $\gamma_{5}$ = 0.1333, respectively. For the first part, we use the Adam optimizer for generators with a learning rate of 0.0002 and SGD optimizer for classifiers with a learning rate of 0.0001 and a batch size of 1. We train first part for 200 epochs keeping the same learning rate for the first 100 epochs and linearly decaying it to zero over the next 100 epochs. Then, we choose the classifier $F'_{S}$ and $G_{ST}$ with the best validation performance during the first stage~(image translation training) and use the classifier and the adapted images $x'_{S}$ generated by $G_{ST}$ during the second stage. For the second part, we use the Adam optimizer with a batch size of 64 and a learning rate of 0.0001. We train $D_{feat}$ and $F'_{S}$ for 200 epochs and $F'_{S}$ is updated only when $D_{feat}$'s accuracy is larger than 0.8. 
% For Twitter-LDL and Flickr-LDL datasets, we divide the datasets by a ratio of 4:1 as training and testing datasets and for ArtPhoto and FI datasets, the ratio is 9:1. 
All of our experiments are conducted on a machine with 4 NVIDIA TITAN $V$ GPUs, each with 12 GB memory.

\begin{table*} 
\centering\footnotesize
\caption{Comparison between the proposed dynamic emotional semantic consistency (DESC) loss and the original ESC loss in~\cite{zhao2019cycleemotiongan} for emotion distribution learning. We use CycleGAN and CycleGAN+Feat as baselines.}
\begin{center} 
\begin{tabular}{c|c|cccccc} 
\toprule
DA setting & Method & $SSD\downarrow$ & $KL\downarrow$ & $BC\uparrow$ & $Canbe\downarrow$ & $Cheb\downarrow$ & $Cos\uparrow$ \\
\hline  
\multirow{4}*{Twitter-LDL $\rightarrow$ Flickr-LDL} & CycleGAN+ESC& 0.1672 & 0.6123 & 0.8156 & 5.8374 & 0.2775 & 0.8089\\
& CycleGAN+DESC & 0.1592 & 0.6027 & 0.8184 & 5.8149 & 0.2732 & 0.8156\\
\hhline{~|-|-|-|-|-|-|-}
%\hline
& CycleGAN+Feat+ESC & 0.1589 & 0.5914 & 0.8130 & 5.7576 & 0.2757 & 0.8126\\
& CycleGAN+Feat+DESC & 0.1558 & 0.5901 & 0.8245 & 5.7745 & 0.2713 & 0.8145\\
\midrule
\multirow{4}*{Flickr-LDL $\rightarrow$ Twitter-LDL} & CycleGAN+ESC& 0.1541 & 0.5800 & 0.8124 & 6.0512 & 0.2688 & 0.8327\\
& CycleGAN+DESC & 0.1478 & 0.5724 & 0.8176 & 6.0701 & 0.2627 & 0.8395 \\
\hhline{~|-|-|-|-|-|-|-}
%\hline
& CycleGAN+Feat+ESC & 0.1493 & 0.5617 & 0.8120 & 6.0178 & 0.2712 & 0.8375\\
& CycleGAN+Feat+DESC & 0.1477 & 0.5513 & 0.8256 & 6.0516 & 0.2593 & 0.8400\\
\bottomrule
\end{tabular}  
\end{center} 
\label{tab:ablation5} 
\end{table*}

\begin{table*}  
\centering\footnotesize
\caption{Comparison between the proposed DESC loss and the original ESC loss in~\cite{zhao2019cycleemotiongan} for dominant emotion classification.}
\begin{center}  
\begin{tabular}{c|c|cccccccc|c} 
\toprule
DA setting & Method & Amu & Ang & Awe & Con & Dis & Exc & Fea & Sad & Avg\\
\hline 
\multirow{4}*{ArtPhoto $\rightarrow$ FI} & CycleGAN+ESC& 55.14 & 12.96 & 36.50 & 4.73 & 6.19 & 4.96 & 1.00 & 63.40 & 27.50 \\
& CycleGAN+DESC & 46.95 & 26.62 & 32.48 & 33.69 & 38.65 & 31.23 & 8.00 & 9.54 & 27.94\\
\hhline{~|-|-|-|-|-|-|-|-|-|-}
%\hline
& CycleGAN+Feat+ESC & 33.74 & 18.90 & 22.83 & 44.27 & 23.31 & 10.18 & 14.00 & 34.63 & 29.62\\
& CycleGAN+Feat+DESC & 2.64 & 21.26 & 38.54 & 37.10 & 12.27 & 23.51 & 11.00 & 47.35 & 30.19\\
\midrule
\multirow{4}*{FI $\rightarrow$ ArtPhoto} & CycleGAN+ESC & 25.00 & 28.57 & 35.00 & 42.86 & 0.00 & 55.00 & 54.55 & 42.42 & 37.34 \\
& CycleGAN+DESC & 30.00 & 7.14 & 35.00 & 35.71 & 35.71 & 55.00 & 36.36 & 51.52 & 37.97\\
\hhline{~|-|-|-|-|-|-|-|-|-|-}
%\hline
& CycleGAN+Feat+ESC & 20.00 & 21.43 & 45.00 & 0.00 & 35.71 & 40.00 & 59.09 & 57.58 & 38.61\\
& CycleGAN+Feat+DESC & 25.00 & 14.29 & 50.00 & 35.71 & 7.14 & 40.00 & 50.00 & 57.58 & 39.24\\
\bottomrule
\end{tabular}  
\end{center} 
\label{tab:ablation7} 
\end{table*}

\subsection{Results and Analysis}
\textbf{Comparison With State-of-the-art.}
The performance comparisons between the proposed CycleEmotionGAN++ model and state-of-the-art approaches are shown in Table~\ref{tab:art-fi} to Table~\ref{tab:fli-twi}. From the results, we have several observations:

(1) The source-only method directly transferring the models trained on the source domain to the target domain performs the worst in all adaptation settings. Due to the influence of \emph{domain shift}, the style of images and distribution of labels are totally different in the two different domains which results in the model’s low transferability from one domain to another.

%All the color transfer (CycleGAN, SAPE, EICT, and TAECT) and domain adaptation methods (ADDA, CyCADA, SimGAN, CycleEmotionGAN, and CycleEmotionGAN++)
(2) All the style transfer and domain adaptation methods outperform the source-only method, with CycleEmotionGAN++ performing the best since these methods can overcome the \emph{domain shift} to some extent. Specifically, the performance improvements of our model over source-only, CycleGAN, SAPE, EICT, TAECT, ADDA, SimGAN, CyCADA measured by KL are 15.67$\%$, 10.39$\%$, 9.46$\%$, 12.73$\%$, 12.40$\%$, 11.05$\%$, 11.27$\%$, and 2.40$\%$ when adapting from the source Twitter-LDL to the target Flickr-LDL, respectively. And the performance improvements of our model over these methods measured by average classification accuracy are 34.16$\%$, 23.16$\%$, 22.93$\%$, 33.38$\%$, 27.58$\%$, 21.57$\%$, 22.50$\%$, and 8.07$\%$ when adapting from the source ArtPhoto to the target FI, respectively. The improvements imply that our model can achieve superior performance relative to state-of-the-art approaches.

(3) For the Twitter-LDL and Flickr-LDL datasets, we observe that our model obtains the best performance in most of the evaluation metrics except BC in the Twitter-LDL$ \rightarrow $Flickr-LDL process and Canbe in the Flickr-LDL$ \rightarrow $Twitter-LDL process. For ArtPhoto and FI datasets, a better model cannot ensure a better accuracy on every single category, but only for the average accuracy. The two methods of (dynamic) emotional semantic consistency loss obtain similar performance. For the original CycleEmotionGAN~\cite{zhao2019cycleemotiongan}, Mikels' wheel obtains better performance, while SKL outperforms Mikels' wheel for CycleEmotionGAN++.
%with the best average classification accuracy among all the methods.

%(4) For some emotion categories, such as \emph{amusement}, \emph{awe} and \emph{sad}, almost all the adaptation methods achieve similar accuracy. For other emotion categories, different methods get quite different performances. 
% Nearly all the adaptation methods get similar classification accuracy in some single emotion catgories such as  while others get totally different accuracy when going from ArtPhoto to FI.

(4) The oracle method achieves the best performance on both emotion distribution learning and dominant emotion classification tasks. However, this model is trained using the ground truth emotion labels from the target domain $\textbf{x}_{T}$, which are of course unavailable in the UDA setting.

(5) For the ArtPhoto and FI datasets, there is still an obvious performance gap between all adaptation methods and the oracle method, especially when adapting from the small-scale ArtPhoto to the large-scale FI. Due to the complexity
and subjectivity of emotions~\cite{yang2018retrieving}, we find the accuracies of all domain adaptation methods are not very high and effectively adapting image emotions is still a challenging problem.

\textbf{Ablation Study.} First, we perform various experiments to evaluate how each component contributes to the adaptation performance, with results shown in  Table~\ref{tab:ablation1} and Table~\ref{tab:ablation3}. We can observe that: \textbf{(1)} Dynamic emotional semantic consistency loss boosts the performance by a large margin; after adding it, the performance improves significantly, proving that preserving the emotion label is of vital importance.
%The performance improvements of +DESC over baseline measured by KL and average classification accuracy are 6.43$\%$, 10.45$\%$, 7.50$\%$, and 19.97$\%$ respectively. 
%We can find the performance improves significantly after adding dynamic emotional semantic consistency loss which proves that preserving the emotion label is of vital importance.
Since we use the classifier trained on adapted domain to test, we should make sure that it can use emotion labels from the source domain as its labels. \textbf{(2)} From the third and fourth rows of each DA setting, we can see the improvements that feature-level loss and multi-scale structural similarity contribute. Each of these two components can improve the performance of the model trained in the source domain. \textbf{(3)} The last row of each setting which contains all of the three components performs the best. For example, it obtains the best average classification accuracy on ArtPhoto and FI datasets.

\begin{figure*}[!t]
    \centering
    \includegraphics[width=0.88\linewidth]{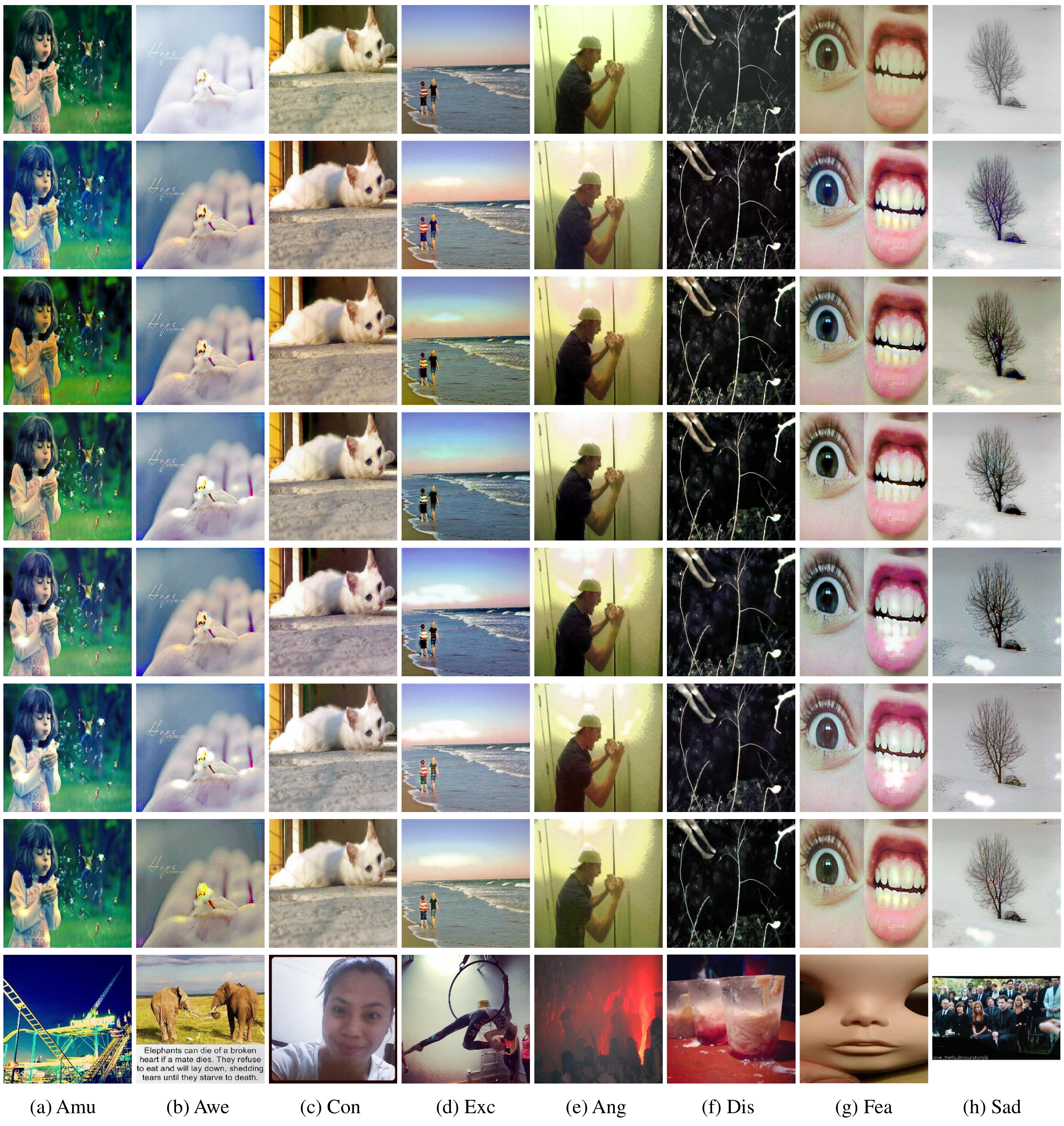}
    \caption{Visualization of images across Mikels' emotion categories from ArtPhoto, adapted in order to make them have the style of FI. From top to bottom are: original ArtPhoto images, images generated by CycleGAN~\cite{zhu2017unpaired}, SAPE~\cite{yan2016automatic}, EICT~\cite{liu2018emotional}, TAECT~\cite{liu2018texture}, CycleEmotionGAN-SKL~\cite{zhao2019cycleemotiongan}, CycleEmotionGAN++-SKL, and original FI images.}
  \label{fig:adaptedart}
\end{figure*}

Second, we compare the proposed dynamic semantic consistency (DESC) loss and the original emotional semantic consistency (ESC) loss. The main difference between these two methods is that DESC uses two classifiers, one for each of the source domain and the adapted domain to dynamically preserve the emotion labels. The results are shown in Table~\ref{tab:ablation5} and Table~\ref{tab:ablation7}. For each process, we use CycleGAN and CycleGAN+Feat as baselines. For Table~\ref{tab:ablation5}, the latter of every two rows obtains better performance in all evaluation metrics except Canbe. For Table~\ref{tab:ablation7}, the latter of every two rows obtains better average classification accuracy.

%As illustrated from Fig.~\ref{fig:adaptedart} and Fig.~\ref{fig:adaptedfi}, 

\textbf{Visualization of Adapted Images.} As illustrated from Fig.~\ref{fig:adaptedart}, we visualize the adapted images to demonstrate the effectiveness and necessity of image translation. We compare the adapted images generated by CycleGAN~\cite{zhu2017unpaired}, SAPE~\cite{yan2016automatic}, EICT~\cite{liu2018emotional}, TAECT~\cite{liu2018texture}, CycleEmotionGAN-SKL, and CycleEmotionGAN++-SKL. We can observe that all these methods can adapt the source images to be more similar to the images of target domain. However, CycleEmotionGAN++-SKL performs better than the other methods. For example, the hue of the adapted image (b) in the second but last line generated by CycleEmotionGAN++-SKL is more yellow which is closer to target domain FI. Therefore, the images generated by our model are more similar to the target images, as compared to the original images and the images generated by CycleGAN. This further demonstrates the effectiveness of the proposed model.

Similar to GAN~\cite{goodfellow2014generative} and CycleGAN~\cite{zhu2017unpaired} based image generation methods, the proposed CycleEmotionGAN++ also suffers from low quality problem. As our goal is to improve the accuracy of task classifier, \textit{i.e.} $F'_{S}$, we did not employ any high-resolution or high-quality generation methods, such as MSG-GAN~\cite{karnewar2020msg} and EventSR~\cite{wang2020eventsr}, which usually require more computation cost. We leave generating high-quality images as our future work.

%The proposed method generates images with similar quality to general GAN. Our goal is to improve the accuracy of the f network. As suggested, we provide some examples of the images generated by CycleGAN and our method, as shown from Table 5 to Table 8.

\begin{figure*}[!t]
    \centering
    \includegraphics[width=0.88\linewidth]{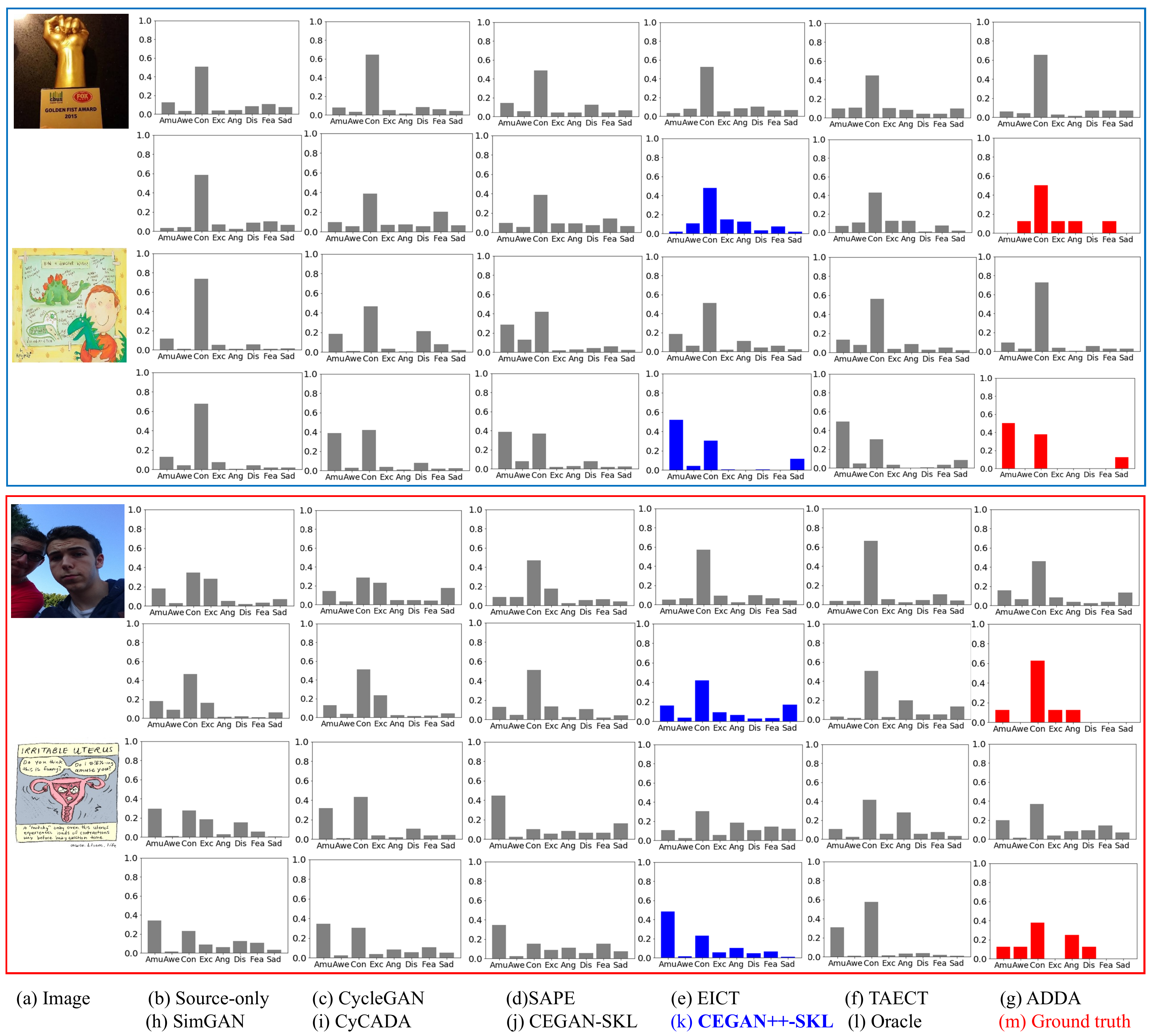}
    \caption{Visualization of predicted emotion distributions on Twitter-LDL using the proposed CycleEmotionGAN++-SKL (CEGAN++-SKL), the original CycleEmotionGAN-SKL (CEGAN-SKL)~\cite{zhao2019cycleemotiongan} and several state-of-the-art approaches (source-only, CycleGAN~\cite{zhu2017unpaired}, SAPE~\cite{yan2016automatic}, EICT~\cite{liu2018emotional}, TAECT~\cite{liu2018texture}, ADDA~\cite{tzeng2017adversarial}, SimGAN~\cite{shrivastava2017learning}, CyCADA~\cite{hoffman2018CyCADA}, and oracle). Original images and the corresponding ground truth distributions are shown in the first column and last image of each group respectively.}
  \label{fig:predictedtwi}
\end{figure*}

\textbf{Visualization of Predicted Results.} Some predicted emotion label distributions are visualized in Fig.~\ref{fig:predictedtwi} for the Twitter-LDL datasets. The first two examples in the red frame show that our model's results are close to the ground truth label distributions, which demonstrates the effectiveness of our proposed model. In other examples in the blue frame, the results of our model, as well as the oracle, are not close enough to the ground truth. In these two examples, we can observe that, though the oracle performs better than all the adaptation methods, it is still very different from the ground truth, demonstrating the need for further research. %The relatively worse results of our model 

%This is reasonable since our model and the oracle performance in all of the metrics is much worse than the ground truth label. %However, no matter whether our model's results are close to ground truth label distributions, the proposed CycleEmotionGAN++ model achieves the highest performance, which demonstrates the effectiveness of the proposed framework. 

% \begin{figure*}[!ht] 
%   \begin{center}
%     \subfigure
%     {
%         \includegraphics[width=.998\linewidth]{./figures/Predicted/twitter1.pdf}
%     }
%     \subfigure
%     {
%         \includegraphics[width=.998\linewidth]{./figures/Predicted/twitter2.pdf}
%     }
%     \end{center}
%   \vfill
%   \caption{Visualization of predicted emotion distributions on Twitter-LDL using the proposed CycleEmotionGAN++-SKL, the original CycleEmotionGAN-SKL~\cite{zhao2019cycleemotiongan} and several state-of-the-art approaches (source-only, CycleGAN~\cite{zhu2017unpaired}, SAPE~\cite{yan2016automatic}, EICT~\cite{liu2018emotional}, TAECT~\cite{liu2018texture}, ADDA~\cite{tzeng2017adversarial}, SimGAN~\cite{shrivastava2017learning}, CyCADA~\cite{hoffman2018CyCADA}, and oracle). Original images and the corresponding ground truth distributions are shown in the first column and the red images of each group respectively.}
%   \label{fig:predictedtwi}
% \end{figure*}

\begin{figure*}[!ht] 
  \centering 
  \begin{minipage}[b]{\linewidth} 
  
  \subfigure[Training/validation (first part) loss]{
    \begin{minipage}[b]{0.235\linewidth} 
      \centering
      \includegraphics[width=1.08\linewidth]{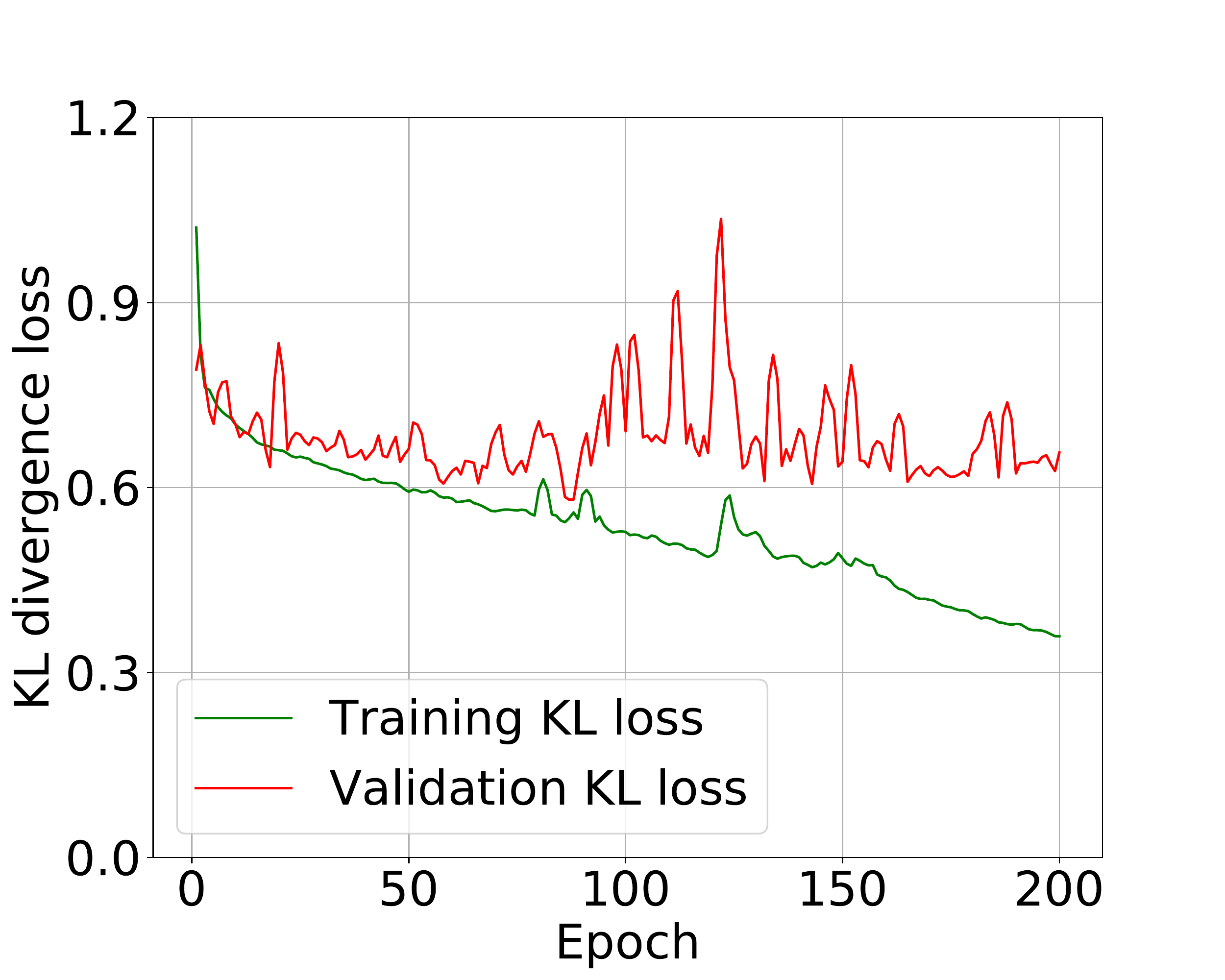}
    \end{minipage}}
  \hfill
  \subfigure[$D_{T}$ and $D_{S}$ GAN loss]{
    \begin{minipage}[b]{0.235\linewidth} 
      \centering
      \includegraphics[width=1.08\linewidth]{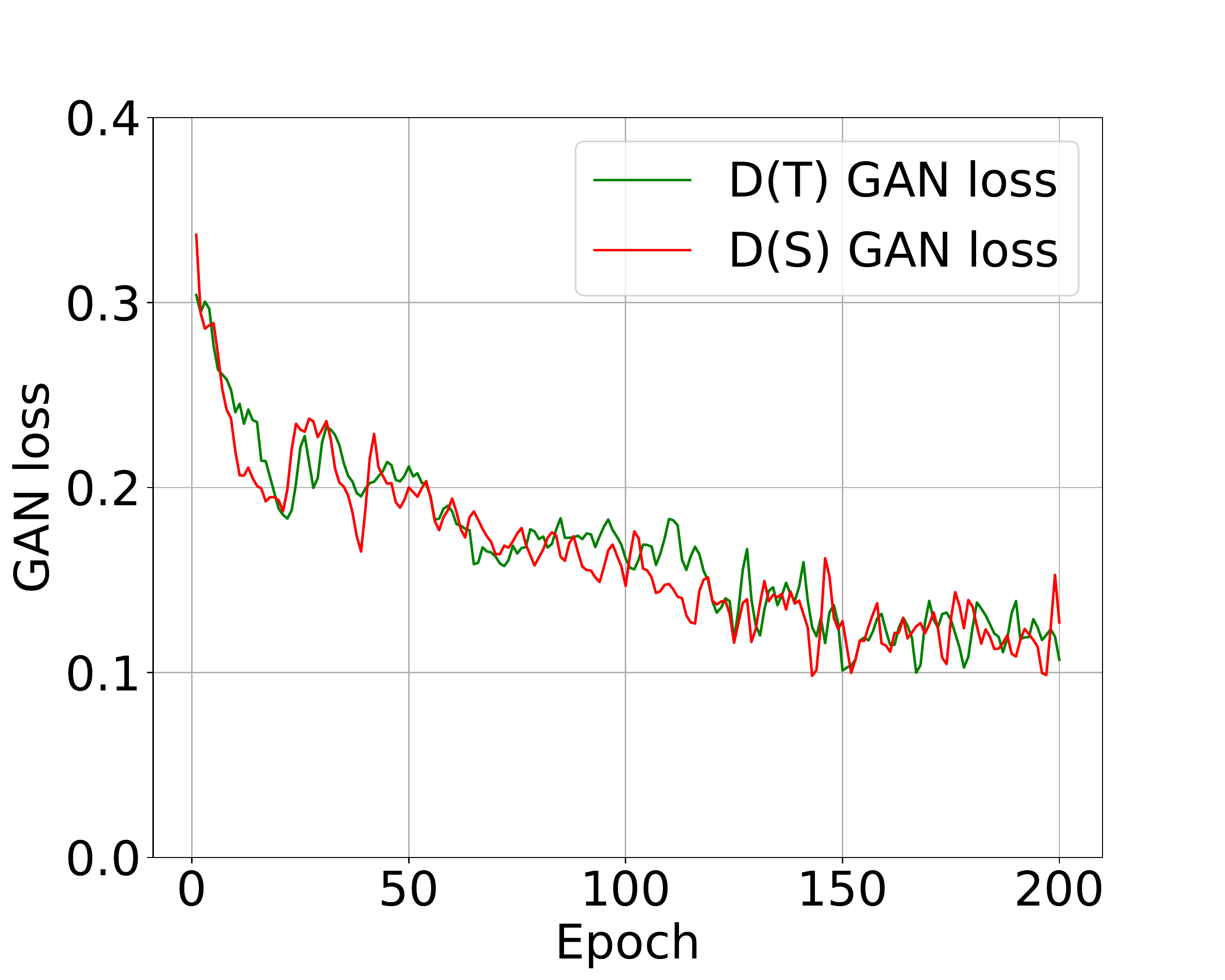}
    \end{minipage}}
  \hfill
  \subfigure[$D_{feat}$ GAN loss]{
    \begin{minipage}[b]{0.235\linewidth} 
      \centering
      \includegraphics[width=1.08\linewidth]{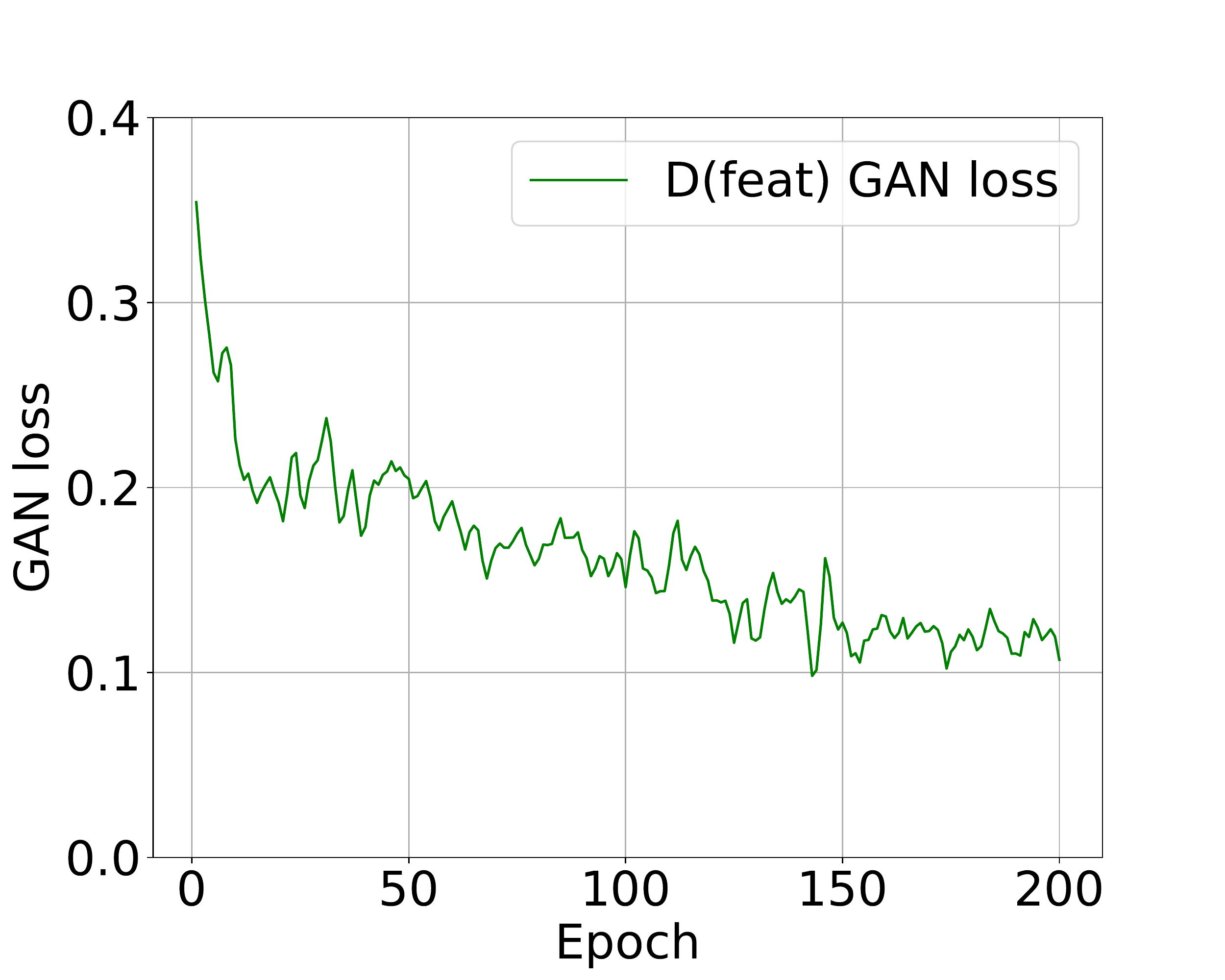}
    \end{minipage}}
  \hfill
  \subfigure[Mixed cycle consistency loss]{
    \begin{minipage}[b]{0.235\linewidth} 
      \centering
      \includegraphics[width=1.08\linewidth]{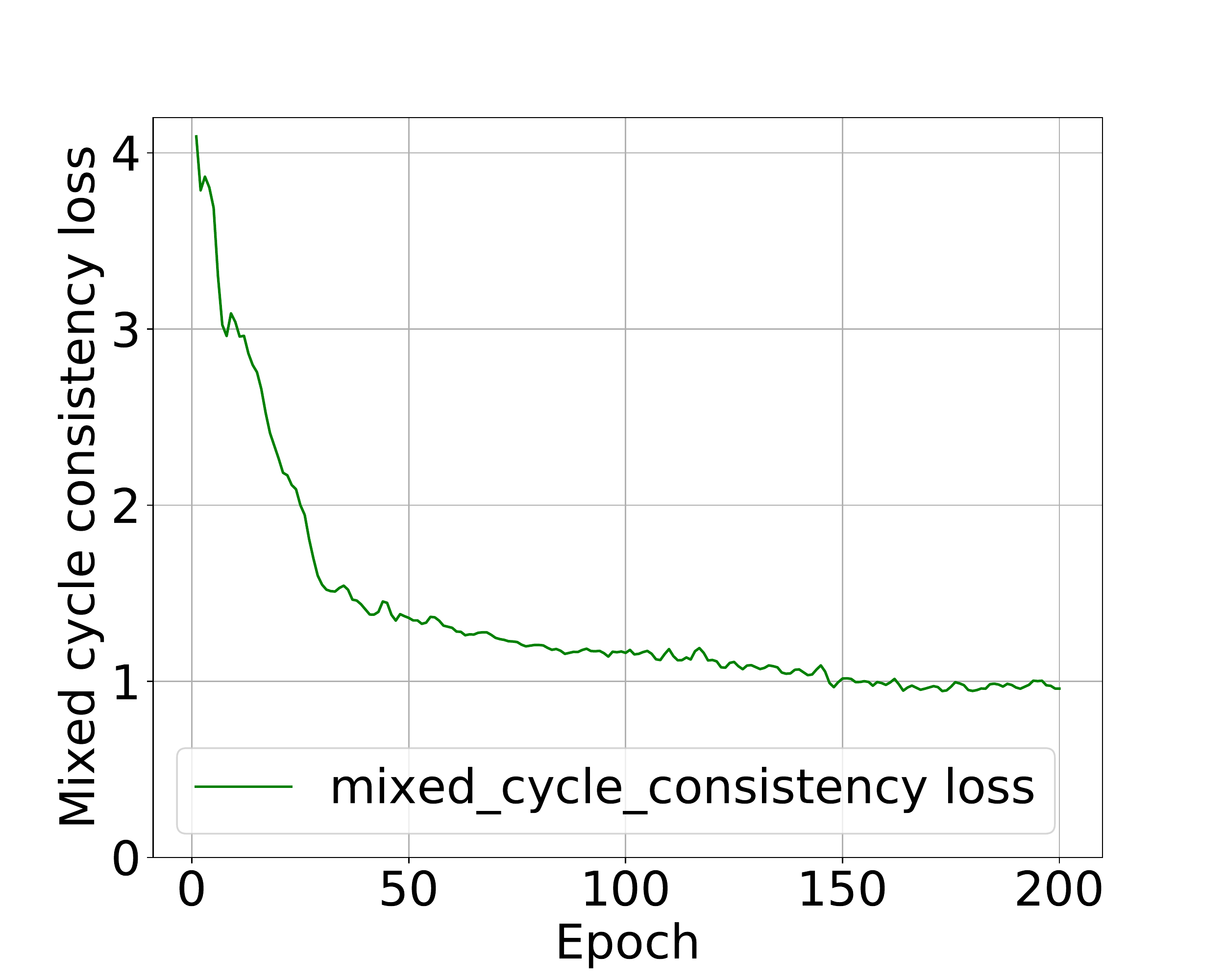}
    \end{minipage}}
    
  \end{minipage}
  \vfill
  \caption{Examples of loss curves when adapting from the source domain Twitter-LDL to the target domain Flickr-LDL. The figures, from left to right show: (a) Training and validation KL divergence loss, (b) GAN loss of $D_{T}$ and $D_{S}$, (c) GAN loss of $D_{feat}$, and (d) mixed cycle consistency loss.}
  \label{fig:curve}
\end{figure*}

\textbf{Convergence.} In order to display the training process more directly, we visualize some loss curves when adapting from the source domain Twitter-LDL to the target domain Flickr-LDL in Fig.~\ref{fig:curve}. We can observe that during the first part training of generating an adapted domain $\{x'_{S}\}$, the best validation KL performance appears between 50 and 100 epochs and then with the decrease of the training loss, the validation KL performance becomes unstable, showing overfitting in Fig.~\ref{fig:curve} (a). We use the networks with best performance, \textit{i.e.} $F'_{S}$ and $G_{ST}$, to generated adapted images $x'_{S}$ for the second part training. From the other three figures, we observe that the losses decrease gradually with the increase of epoch number.

\section{Conclusion}
In this paper, we studied the unsupervised domain adaptation (UDA) problem for both emotion distribution learning and dominant emotion classification. We proposed an end-to-end cycle-consistent adversarial model, CycleEmotionGAN++, to bridge the gap between different domains. We generated an adapted domain to align the source and target domains on the pixel-level by improving CycleGAN with a multi-scale structured cycle-consistency loss. During the image translation, we proposed dynamic emotional semantic consistency loss to preserve the emotion labels of the source images. We trained a transferable task classifier on the adapted domain with feature-level alignment between the adapted and target domains. We conducted extensive experiments on the Flickr-LDL and Twitter-LDL datasets for emotion distribution learning, and the ArtPhoto and FI datasets for dominant emotion classification. The results on these four datasets demonstrate the significant improvements yielded by the proposed method over state-of-the-art UDA approaches.
For future work, we plan to extend the CycleEmotionGAN++ model to multi-modal settings, such as audio-visual emotion recognition. We will also investigate domain generalization without accessing target data for VEA and study theoretical deduction to better understand the learning process.

%We will also investigate multi-source domain adaptation with multiple source domains and domain generalization without accessing target data for VEA.

%\section{ACKNOWLEDGEMENTS}
%\label{Ackonwledgements}
%This work was supported by the National Natural Science Foundation of China (No. 61472103) and Key Program (No. 61133003).

\ifCLASSOPTIONcaptionsoff
  \newpage
\fi

\bibliographystyle{IEEEtranN}\small
\bibliography{Refe}

\vfill\break

\end{document}